\documentclass{article} 
\PassOptionsToPackage{table}{xcolor} 
\usepackage{iclr2027_conference,times}

\usepackage{amsmath,amsfonts,bm}

\def\eqref#1{equation~\ref{#1}}

\def\1{\bm{1}}

\DeclareMathAlphabet{\mathsfit}{\encodingdefault}{\sfdefault}{m}{sl}
\SetMathAlphabet{\mathsfit}{bold}{\encodingdefault}{\sfdefault}{bx}{n}

\usepackage{hyperref}
\usepackage{url}
\usepackage{booktabs}
\usepackage{multirow}
\usepackage{graphicx}
\usepackage{capt-of}
\usepackage[table]{xcolor}
\hypersetup{
  colorlinks=true,
  citecolor=green,
  linkcolor=red,
  urlcolor=black
}
\usepackage{array}
\usepackage{tabularx}
\usepackage{pifont}
\definecolor{mygray}{gray}{0.95}
\definecolor{gainBlue}{HTML}{DBE7F8}
\definecolor{comparisonYes}{HTML}{009600}
\definecolor{comparisonNo}{HTML}{BF0000}

\newcommand{\name}{Maco-RL}

\title{\name{}: Scaling Agent Capabilities with Mixed-Task Reinforcement Learning}

\author{
\hspace*{1em}
Jiacheng Zhang$^{1,2}$\thanks{Equal contribution.}~, ~Qixing Zhou$^{1,2}$\footnotemark[1]~, ~Haiyang Wang$^{2}$\thanks{Corresponding Author.}~, ~Qipeng Gu$^{2}$, ~Shuzhe Wu$^{2}$\\
\makebox[\textwidth][c]{
\hspace{-1.7em}\textbf{Feiyang Pan$^{2}$~, ~Lue Fan$^{1}$\footnotemark[2]~~, ~Dandan Tu$^{2}$\footnotemark[2]~, ~Zhaoxiang Zhang$^{1}$\footnotemark[2]}
}\\[2pt]
\makebox[\textwidth][c]{
\hspace{-1.7em}$^1$ Institute of Automation, Chinese Academy of Sciences \quad
$^2$ Huawei Technologies Co., Ltd
}\\[2pt]
\makebox[\textwidth][c]{\tt \hspace{-1.5em}haiyang.wang@huawei.com ~ lue.fan@ia.ac.cn
}
}

\iclrfinalcopy 
\begin{document}

\maketitle
\lhead{}

\vspace{-10pt}
\noindent\begin{minipage}{\linewidth}
  \centering
  \includegraphics[trim=0 39bp 0 0,clip,width=\linewidth]{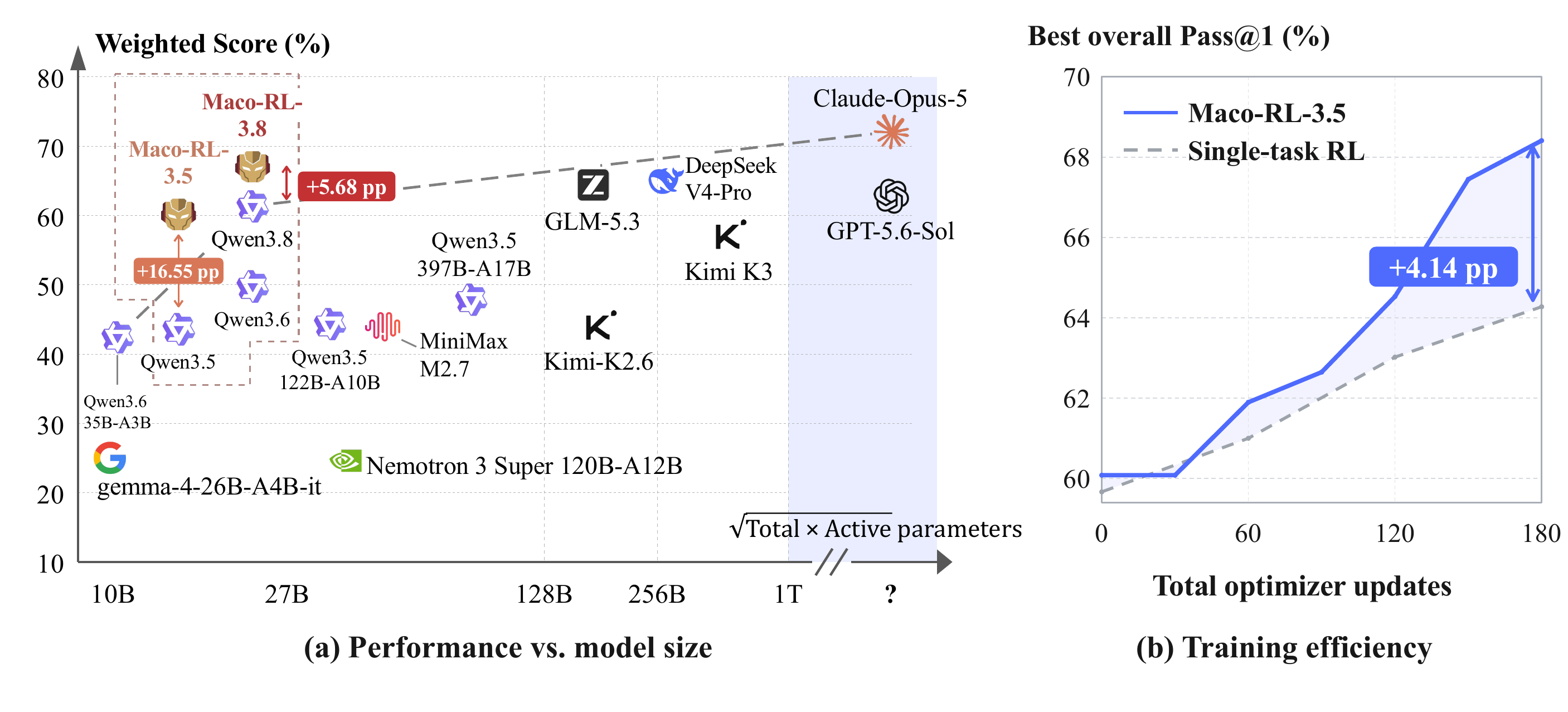}\par
  \captionof{figure}{Our \name~enables 27B-sized models to substantially outperform baselines and even be competitive with much larger models in terms of overall pass rates across multiple task domains, i.e., issue- and feature-oriented coding, terminal and Claw. Experimental results reveal that \name~more effectively turns compute budget into model capability. (a) Average pass rates over five benchmarks, i.e., Pass@1 on SWE-bench Verified, Terminal-Bench 2.1 and Claw-Anything, Tests Passed on FeatureBench 1.1 and $\mathrm{Pass}^{3}$ on Claw-Eval, against model sizes, which are $\sqrt{P_{\mathrm{total}}P_{\mathrm{active}}}$ for MoE models and $P_{\mathrm{total}}$ otherwise. (b) The mean Pass@1 metrics against compute budget for RL on mixed and single tasks. For single-task RL, we sum compute across the three tasks.}
  \label{fig:cover-performance}
\end{minipage}
\par\medskip

\begin{abstract}
Mixed-task reinforcement learning (MixRL) directly trains a unified policy across domains to continually expand agent capabilities without training separate specialists or requiring an extra distillation stage. Despite its appeal, scaling MixRL poses significant challenges in achieving stable and efficient mixed-task learning, effectively allocating rollout compute, and supporting continual domain expansion. To address these challenges, we propose \textit{\name}, an approach that effectively scales MixRL. Specifically, \name{} employs a quota-based scheduler for asynchronous rollouts across multiple task domains, thereby stabilizing training by balancing gradient contributions and mitigating off-policy effects. Along an orthogonal axis, \name{} further introduces Tri-Buffer Sampling (TBS) to allocate compute among samples with varying levels of mastery, promoting sustained frontier exploration while preserving acquired capabilities. Experiments on five benchmarks show \name{} raises Qwen3.5-27B's overall score from 43.6 to 54.2 and then to 60.1 with continual expansion, reaching the frontier held by Qwen3.8-27B, which is further pushed outward by $+5.7$. \name{} also outperforms single-task RL with less compute for policy updates. These results support MixRL as a practical and compute-efficient post-training approach with substantial potential to strengthen general agent capabilities.
\end{abstract}

\section{Introduction}
\label{sec:introduction}
\begin{figure}[t]
  \centering
  \includegraphics[trim=0 13bp 0 0,clip,width=\linewidth]{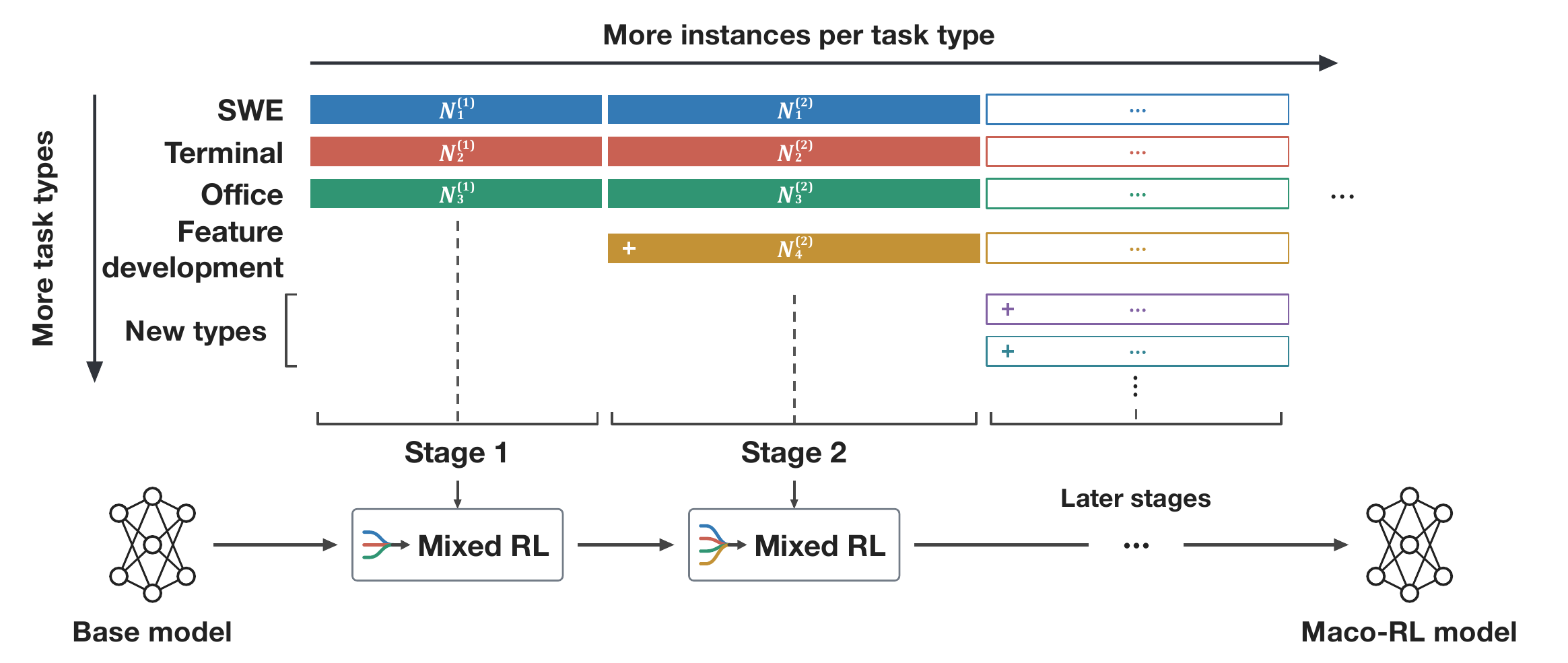}\par
  \vspace{-4pt}
  \caption{Continual mixed RL expands instances per task type horizontally and adds task types vertically. $N_i^{(s)}$ denotes the instance count for task type $i$ in stage $s$.}
  \label{fig:incremental-training}
\end{figure}

Large language model (LLM) agents have made substantial progress in software engineering, terminal operations, and office automation~\citep{jimenez2024swe,merrill2026terminalbench,ye2026claweval}. Reinforcement learning (RL) has played a key role in these advances by enabling agents to learn from environment feedback during post-training~\citep{luo2025deepswe,ivison2026tmax,song2026clawgym,zhang2025agentrl}. This progress motivates us to explore an RL-only path towards more capable general-purpose agents.

Several recent agent post-training pipelines optimize domain specialists independently with RL, then consolidate their capabilities through on-policy distillation (OPD)~\citep{deepseek2026v4,kimi2026k3,glm2026glm5,xiaomi2026mimov2flash}. Another promising direction is direct mixed-task RL (MixRL), which jointly optimizes a shared policy across multiple domains~\citep{zhang2025agentrl,xi2026generalization}. By allowing tasks to benefit from shared training experience, MixRL offers the potential for better performance and learning efficiency without a separate stage for consolidating specialist capabilities. We therefore study how to scale direct MixRL to larger and more diverse collections of tasks and environments.

However, realizing the benefits of MixRL at scale poses several technical challenges. Highly heterogeneous tasks can differ markedly in their reward distributions and potentially induce conflicting optimization directions~\citep{yu2020gradient,zhang2025agentrl}, limiting performance gains from joint training. Stable MixRL training also places substantial demands on infrastructure, requiring close coordination between rollout generation and policy updates across diverse environments~\citep{sheng2025hybridflow,lai2026computerrl}. Beyond joint optimization, the high computational cost of agentic RL makes effective task sampling essential~\citep{yu2025dapo}. The sampling strategy must balance reinforcing occasional successes to establish reliable performance with exploring unsolved instances to expand the policy's capabilities. During training runs spanning weeks or months in practice, new tasks and data may also become available for further scaling. Efficient scaling therefore requires incorporating these additions incrementally while preserving previously acquired capabilities.

To this end, we introduce \name{}, a framework for scaling direct mixed-task agentic RL. Each policy update mixes gradients from trajectories collected across multiple tasks, supported by an efficient asynchronous rollout--training pipeline that accommodates widely varying rollout costs. Within this pipeline, we propose Tri-Buffer Sampling (TBS) to organize instance sampling through three buffers with separate quotas. The Frontier Buffer sustains exploration by reserving sampling capacity for instances the policy has yet to solve. The Reinforcement Buffer uses recent success rates to concentrate rollout resources on occasionally solved instances, consolidating emerging capabilities. The History Buffer replays instances from earlier training stages with fresh rollouts from the current policy, supporting progressive scaling while helping retain acquired capabilities.

As illustrated in Figure~\ref{fig:incremental-training}, the overall MixRL scaling process comprises two stages. We first jointly train on code repair, terminal operations, and office automation, then expand the existing task pools and introduce feature development as a fourth task type. With RL-only post-training, \name{} enables Qwen3.5-27B to achieve overall performance comparable to that of Qwen3.8-27B across five public agentic benchmarks spanning three domains. Applied to the stronger Qwen3.8-27B model, \name{} achieves the highest overall score among the evaluated 27B models, even surpassing several substantially larger models. Compared with single-task RL, MixRL also achieves better performance with fewer cumulative policy updates (Figure~\ref{fig:cover-performance}). Ablation studies show that TBS improves learning efficiency while supporting the discovery of solutions to previously unsolved instances, and that mixed-task updates stabilize training. Together, these results demonstrate the potential of MixRL-only post-training to advance general agent capabilities.

Our contributions are (1) \name{}, a quota-based asynchronous training framework that combines per-environment rollout concurrency with mixed-task policy updates; (2) TBS, a unified instance-sampling strategy that balances exploration, reinforcement, and task replay during incremental expansion; and (3) cross-model and cross-stage experiments, controlled ablations, and analyses of training failures that inform practical long-horizon LLM-agent training.

\section{Related Work}
\label{sec:related-work}

\paragraph*{Multi-Task Reinforcement Learning}
MTRL studies shared learning across tasks~\citep{varghese2020survey}. Distral shares knowledge through a distilled policy~\citep{teh2017distral}, while \citet{deramo2020sharing} establish theoretical benefits of representation sharing. PopArt balances learning across reward scales~\citep{hessel2019popart}, and IMPALA combines distributed actor--learner training with off-policy correction~\citep{espeholt2018impala}. Building on these foundations, \name{} addresses the stability and scaling challenges of MTRL for long-horizon LLM agents across heterogeneous task environments and expanding task pools.

\paragraph*{Mixed-Task Agentic RL}
AgentTuning~\citep{zeng2024agenttuning} develops shared agent capabilities through multi-task SFT, while AgentGym~\citep{xi2025agentgym} combines demonstrations with cross-environment interaction for reward-weighted learning. AgentRL~\citep{zhang2025agentrl} jointly trains agents across environments using RL. Complementing joint training, \citet{xi2026generalization} study transfer and retention under joint and sequential RL, and ACuRL~\citep{xue2026acurl} uses experience-driven curricula for continual GUI adaptation. Building on this broader interest in cross-environment learning, \name{} focuses on joint RL as task instances and task types expand, coordinating new capability acquisition with existing capability retention.

\paragraph*{Task Sampling for Agentic RL}
Task scheduling and experience prioritization provide related foundations. Scheduled Multi-Task Training prioritizes difficult tasks to counter simplicity bias~\citep{cho2024hard}, while prioritized experience replay revisits stored transitions according to temporal-difference errors~\citep{schaul2016prioritized}. For LLMs, LILO and CDAS guide sampling by outcome variability or estimated competence~\citep{foster2025lilo,kong2026cdas}. Interactive approaches use value-guided data selection, failure-driven curricula, adaptive rollout counts, or success-rate-based sampling~\citep{bai2024digirl,qi2025webrl,li2025dart,lv2026scalecua,huang2026evocua}. TBS combines related principles in a unified allocation policy: it reserves separate quotas for unsolved, previously solved, and earlier-stage instances, then favors occasional successes within the Reinforcement Buffer. Unlike transition replay, it samples task instances to generate fresh rollouts. Its quotas prevent a large unsolved pool from dominating rollout allocation while preserving exploration and task replay as new data and domains are introduced.

\paragraph*{Training Frameworks for Agentic RL}
Agentic RL requires efficient interaction and stable optimization. RAGEN~\citep{wang2025ragen} studies reward variability and gradient instability, while AgentRL~\citep{zhang2025agentrl} combines asynchronous training with task-level advantage normalization, and ComputerRL~\citep{lai2026computerrl} scales online RL through parallel environments. General-purpose frameworks provide complementary support: verl~\citep{sheng2025hybridflow} uses hybrid control and model resharding for efficient execution, while slime~\citep{slimeteam2025slime} integrates Megatron-LM and SGLang with customizable rollout interfaces and asynchronous execution. To accommodate heterogeneous interaction times and task difficulties, \name{} combines asynchronous training and per-environment resource isolation with gradient mixing for efficient and stable updates.

\begin{figure}[htbp]
  \centering
  \includegraphics[width=\linewidth]{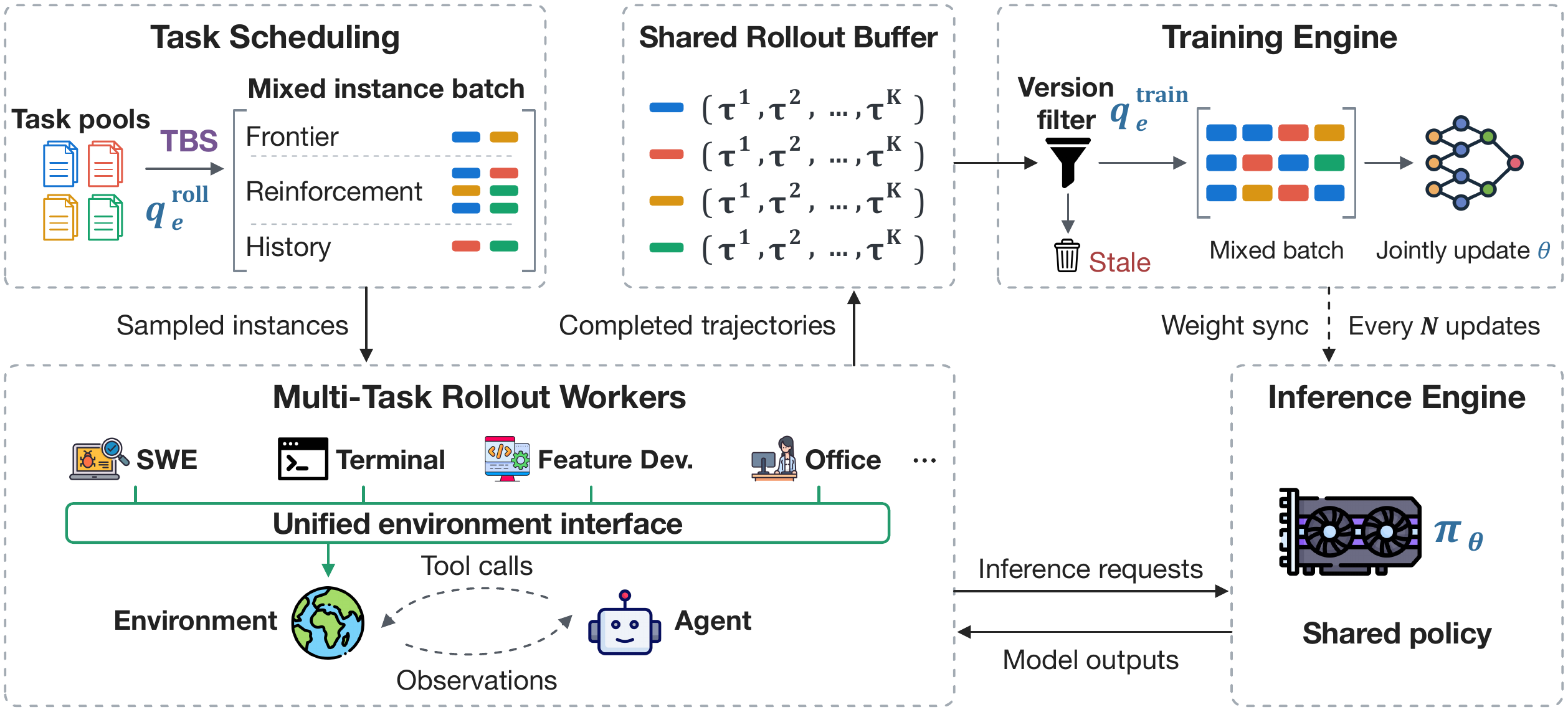}
  \caption{Overview of \name{}. TBS samples instances from each task pool. Rollouts run asynchronously across task types and in parallel with training. A shared rollout buffer collects completed trajectories for joint policy updates on mixed batches.}
  \label{fig:expanserl-framework}
\end{figure}

\section{Methodology}
\label{sec:method}
\name{} supports direct MixRL across diverse tasks, with incremental additions of task instances and task types (Section~\ref{sec:training-overview}). It combines asynchronous rollout generation with gradient mixing for efficient and stable joint training (Section~\ref{sec:training-framework}). Tri-Buffer Sampling (TBS) allocates the sampling budget to balance exploration of unsolved instances, reinforcement of emerging capabilities, and retention of existing capabilities (Section~\ref{sec:fsp}).

\subsection{Tasks and Progressive RL Scaling}
\label{sec:training-overview}
\name{} trains a shared policy on four task types. Code repair, terminal operations, and office automation follow the task settings exemplified by SWE-bench, Terminal-Bench, and Claw style, respectively. Feature development requires substantial coding across multiple files to implement complete features from product requirements documents.

In practice, agentic RL training typically spans weeks or months, during which additional data and new task types become available and need to be incorporated into ongoing training. To simulate this process, we adopt a two-stage incremental training schedule, as illustrated in Figure~\ref{fig:incremental-training}. Stage 1 jointly trains on the first three task types. Stage 2 continues from the checkpoint, expanding the pools for these tasks and introducing feature development. More details are in Appendix \ref{app:environments}.

\subsection{Asynchronous Mixed-Task Training Framework}
\label{sec:training-framework}
\name{} decouples rollout generation from policy optimization to support joint training across heterogeneous tasks (Figure~\ref{fig:expanserl-framework}). Rollout workers independently sample instances from their respective task pools and collect completed trajectories in a shared rollout buffer. 
The learner assembles mixed-task batches from this buffer using fixed per-task instance quotas, ensuring that all task types are included in every update. To limit data staleness, we periodically synchronize the rollout policy with the learner and discard trajectories whose policy-version lag exceeds a predefined threshold.
\paragraph{Rollout allocation for efficient batch assembly.}
Because each batch must satisfy all task quotas, tasks with longer rollout durations can delay policy updates even when other tasks have sufficient trajectories. We therefore allocate more concurrent rollout slots to slower tasks to align trajectory supply with batch requirements and reduce learner idle time.
\paragraph{Mixing tasks within each batch.} MixRL can either alternate single-task batches or combine trajectories from all task types within each batch. We adopt the latter, arguing that controlling policy drift is more critical under asynchronous execution than mitigating conflicts between task objectives. Delayed weight synchronization causes the rollout policy to lag behind the current training policy, and task alternation may further widen this shift. Our ablation shows tool-use failures and reward collapse under task alternation, whereas training on mixed batches remains stable (Figure~\ref{fig:gradient-mixing-reward}).

By coordinating asynchronous rollouts with policy updates on mixed batches, \name{} maintains high training throughput and stable optimization, supporting continued gains in agent performance. Appendix~\ref{app:async-pipeline} provides implementation details and training pseudocode (Figure~\ref{fig:app-async-pseudocode}).

\subsection{Tri-Buffer Sampling: Balancing Future, Present, and Past}
\label{sec:fsp}
\begin{figure}[t]
  \centering
  \includegraphics[width=\linewidth]{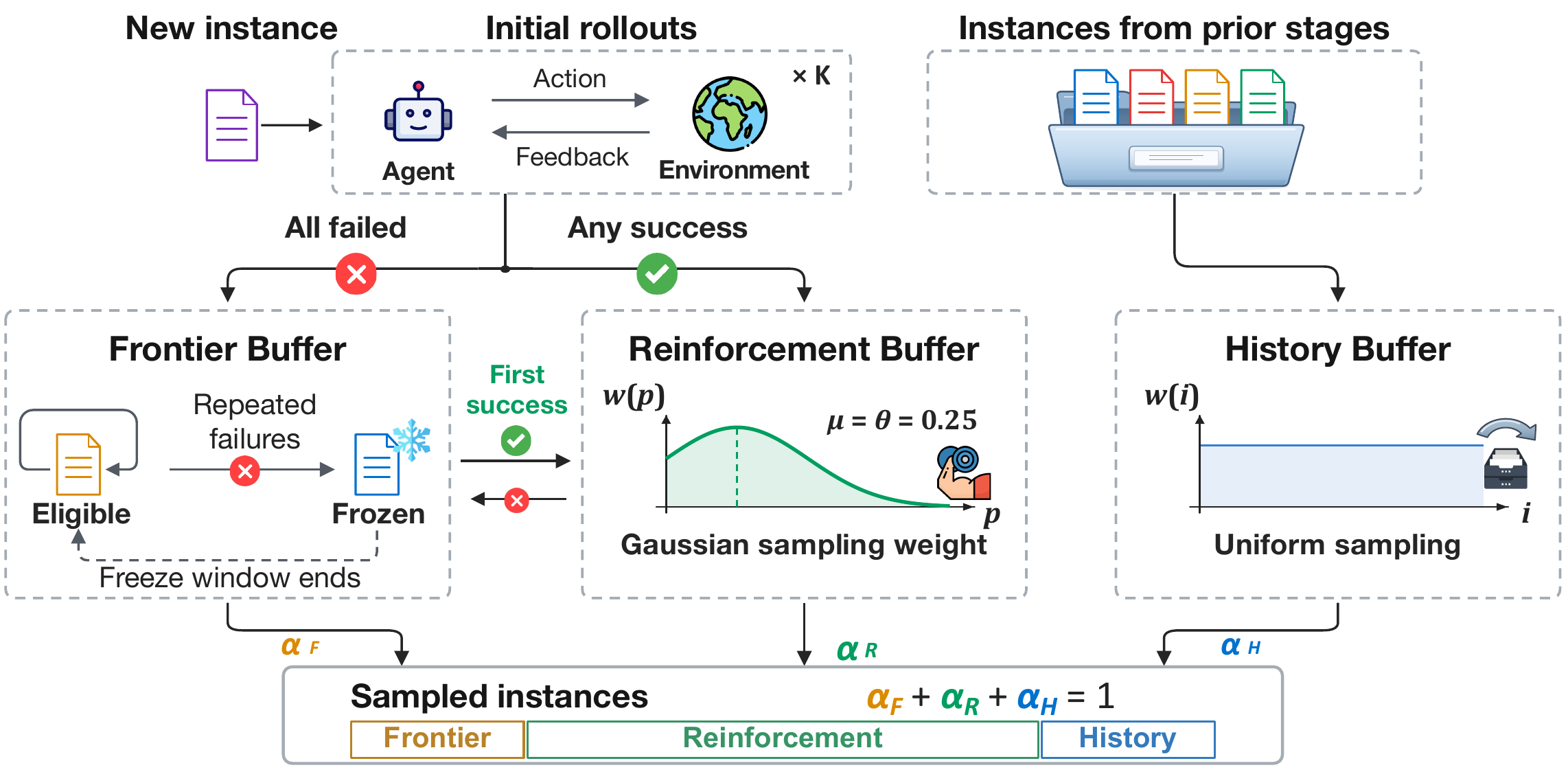}
  \caption{Overview of Tri-Buffer Sampling (TBS). Frontier, Reinforcement, and History buffers receive fixed sampling quotas for exploring unsolved instances, consolidating emerging capabilities, and replaying instances from earlier stages, respectively.}
  \label{fig:frontier-sampling-policy}
\end{figure}

TBS is a unified instance-sampling strategy for exploration, reinforcement, and task replay during incremental RL scaling. In sparse-reward tasks, rollout groups with identical zero rewards yield zero GRPO-style advantages, so repeatedly sampling unsolved instances can consume substantial compute without a reward-based policy-gradient signal. TBS allocates target proportions $\alpha_{\mathrm{F}}$, $\alpha_{\mathrm{R}}$, and $\alpha_{\mathrm{H}}$ to Frontier, Reinforcement, and History buffers, with $\alpha_{\mathrm{F}}+\alpha_{\mathrm{R}}+\alpha_{\mathrm{H}}=1$ (Figure~\ref{fig:frontier-sampling-policy}). These quotas explicitly balance the three objectives rather than allowing their allocation to follow the data distribution. We evaluate each newly introduced instance with an initial group of rollouts. Instances with no successful rollout enter the Frontier Buffer, while those with at least one success enter the Reinforcement Buffer.

\paragraph{Frontier Buffer: Exploring Unsolved Instances.}
The Frontier Buffer reserves sampling capacity for instances with no observed success. To limit unproductive retries, we temporarily suspend sampling of instances that remain unsuccessful across consecutive rollout groups and restore their eligibility after a predefined cooldown period. After its first successful rollout, an instance moves to the Reinforcement Buffer and does not return to the Frontier Buffer.

\paragraph{Reinforcement Buffer: Consolidating Emerging Capabilities.}
The Reinforcement Buffer prioritizes instances that the policy has solved but cannot yet solve reliably. For an instance $x$ with success rate $\widehat{p}(x)$ in its latest rollout group, we define its sampling weight and probability as
\begin{equation}
  w(x)=\exp\!\left[-\frac{(\widehat{p}(x)-\mu)^2}{2\theta^2}\right],
  \qquad
  P_{\mathrm{R}}(x)=\frac{w(x)}{\sum_{y\in\mathcal{R}}w(y)},
  \label{eq:reinforcement-sampling}
\end{equation}
Here, $\mathcal{R}$ is the Reinforcement Buffer for the corresponding task type. We set $\mu=\theta=0.25$, favoring low but nonzero recent success rates aiming to turn occasional successful solutions into reliable capabilities.

\paragraph{History Buffer: Retaining Prior Capabilities.} This buffer retains instances from earlier training stages to maintain exposure to past tasks. We sample these instances uniformly and generate fresh rollouts with the current policy, supporting capability retention during progressive RL scaling.

\section{Experiments}
\label{sec:experiments}

\begin{table}[t]
  \centering
  \caption{Main results across five benchmarks (\%). S1 and S2 denote the two training stages of Qwen3.5-27B; S2 continues training from S1 with feature development tasks added. $\dagger$: Terminal-Bench 2.1 scores from \citet{artificialanalysis2026terminalbench21}; $\ddagger$: SWE-bench Verified scores from \citet{valsai2026swebench}. Unmarked benchmark scores are from our evaluations.}
  \label{tab:main-results}
  \footnotesize
  \setlength{\tabcolsep}{0.6pt}
  \setlength{\aboverulesep}{0pt}
  \setlength{\belowrulesep}{0pt}
  \setlength{\extrarowheight}{1.5pt}
  \renewcommand{\arraystretch}{1.12}
    \begin{tabularx}{\linewidth}{@{}>{\raggedright\arraybackslash}Xccccc<{\hspace{5pt}}c>{\columncolor[gray]{0.92}}c@{}}
      \toprule
      \rule{0pt}{20pt}\multirow{2}{*}{Model}
        & \shortstack{Claw-\\Eval}
        & \shortstack{Claw-\\Anything}
        & \shortstack{Terminal-\\Bench 2.1}
        & \shortstack{SWE-bench\\Verified}
        & \multicolumn{2}{c}{\shortstack{FeatureBench 1.1\\(fast split)}}
        & \cellcolor{white} \\
      \cmidrule(lr){6-7}
        & $\mathrm{Pass}^{3}$ & Pass@1 & Pass@1 & Pass@1
        & Resolved & Tests Passed & \cellcolor{white}\multirow{-2}{*}{Overall} \\
      \midrule
      \multicolumn{7}{@{}l}{\textit{Frontier Models}} & \\
        Claude-Opus-5 (high) & 73.9 & 34.7 & 87.6$^{\dagger}$ & 97.0$^{\ddagger}$ & 40.0 & 67.2 & 72.1 \\
        DeepSeek-V4-Pro (high) & 67.7 & 43.3 & 78.7$^{\dagger}$ & 96.4$^{\ddagger}$ & 11.0 & 38.9 & 65.0 \\
        GLM-5.3 (max) & 66.5 & 30.0 & 83.9$^{\dagger}$ & 95.4$^{\ddagger}$ & 12.0 & 46.3 & 64.4 \\
        GPT-5.6-Sol (medium) & 60.3 & 21.3 & 86.1$^{\dagger}$ & 96.2$^{\ddagger}$ & 12.0 & 49.8 & 62.8 \\
        Kimi-K3 (max) & 55.3 & 28.0 & 85.0$^{\dagger}$ & 93.4$^{\ddagger}$ & 7.0 & 22.9 & 56.9 \\
      \midrule
      \multicolumn{7}{@{}l}{\textit{Baseline Models}} & \\
        Qwen3.5-27B & 60.2 & 12.0 & 40.5 & 67.4 & 3.0 & 37.9 & 43.6 \\
        Qwen3.6-27B & 68.3 & 14.2 & 56.2 & 71.2 & 4.0 & 39.0 & 49.8 \\
        Qwen3.8-27B & 71.4 & 46.0 & 68.9 & 70.2 & 9.0 & 50.3 & 61.4 \\
      \midrule
      \rowcolor{mygray}
        \textbf{\name{}-3.5 (S1)} & 71.4 & 51.3 & 52.4 & 67.0 & 6.0 & 28.6 & \cellcolor[gray]{0.92}54.2 \\
      \rowcolor{mygray}
        \textbf{\name{}-3.5 (S2)} & 73.3 & 64.6 & 54.3 & 68.6 & 6.0 & 39.9 & \cellcolor[gray]{0.92}60.1 \\
      \rowcolor{gainBlue}
      \textit{S2 gain over Qwen3.5-27B} & \textbf{+13.1} & \textbf{+52.6} & \textbf{+13.9} & \textbf{+1.2} & \textbf{+3.0} & \textbf{+2.0} & \cellcolor{gainBlue}\textbf{+16.6} \\
      \rowcolor{mygray}
        \textbf{\name{}-3.8} & 74.5 & 55.2 & 72.3 & 79.0 & 12.0 & 54.2 & \cellcolor[gray]{0.92}67.0 \\
      \rowcolor{gainBlue}
      \textit{Gain over Qwen3.8-27B} & \textbf{+3.1} & \textbf{+9.2} & \textbf{+3.4} & \textbf{+8.8} & \textbf{+3.0} & \textbf{+3.9} & \cellcolor{gainBlue}\textbf{+5.7} \\
      \bottomrule
    \end{tabularx}
\end{table}

\subsection{Experimental Setup}
\label{sec:setup}

\paragraph{Training.}
We progressively scale RL-only post-training of Qwen3.5-27B~\citep{qwen2026qwen35} across two stages, incorporating new tasks and data as they are developed. In the first stage (S1), we train on publicly available issue-resolution tasks (SWE), Terminal-Bench-style tasks from TMax~\citep{ivison2026tmax}, and Claw-style tasks from Claw-Anything~\citep{lin2026clawanything}. In the second stage (S2), we add SWE tasks, build more TMax environments, and generate new Claw-Anything tasks. We also introduce a new task type, FeatureBench-style feature development, using CodeWorld, our agent-based adaptation of the FeatureBench~\citep{zhou2026featurebench} data construction pipeline. Table~\ref{tab:training-stages} summarizes instance counts and rollout resource allocations by stage. To assess whether the same training recipe can further improve a stronger base model, we train Qwen3.8-27B in a single stage with all four task types available from the start.

Our RL training builds on verl~\citep{sheng2025hybridflow} and uses RLOO~\citep{ahmadian2024rloo} for advantage estimation, with 8 rollouts per task instance and 256 trajectories per policy update. TBS is applied to the SWE-bench, Terminal-Bench, and FeatureBench task families, while Claw-style tasks are sampled uniformly. Claw uses uniform sampling because its partial-credit rewards can provide within-group reward variation even below the success threshold (Appendix~\ref{app:environments}). The Frontier, Reinforcement, and History buffers receive 20\%, 60\%, and 20\% of the sampling budget, respectively, with $\mu=\theta=0.25$ for Reinforcement Buffer sampling. Environments run in Kubernetes-managed Docker containers. Appendices~\ref{app:environments} and~\ref{app:training} provide data preparation and training details.

\paragraph{Evaluation protocol.}
We compare our models against Qwen baselines and frontier models on five benchmarks. Frontier-model scores on Terminal-Bench 2.1 and SWE-bench Verified are taken from Artificial Analysis and Vals AI, respectively, as marked in Table~\ref{tab:main-results}; all other benchmark scores are from our evaluations. For our evaluations, we use an adapted SWE-Agent~\citep{yang2024sweagent} for SWE-bench Verified~\citep{chowdhury2024sweverified} and FeatureBench 1.1~\citep{zhou2026featurebench}, Terminus-2~\citep{merrill2026terminalbench} for Terminal-Bench 2.1~\citep{terminalbench2026v21}, and the official harnesses for Claw-Eval~\citep{ye2026claweval} and Claw-Anything~\citep{lin2026clawanything}. The Overall score is the unweighted mean of $\mathrm{Pass}^{3}$ on Claw-Eval, Tests Passed on FeatureBench, and Pass@1 on each of the other three benchmarks. Appendix~\ref{app:evaluation} details the evaluation settings.

\begin{table}[t]
  \centering
  \caption{Task instance counts and rollout resource allocations. S1 and Qwen3.8-27B report full training pools; S2 reports only new instances. S2 combines 10,452 new instances with all 11,086 S1 instances retained in the History Buffer, totaling 21,538 instances.}
  \label{tab:training-stages}
  \footnotesize
  \setlength{\tabcolsep}{3pt}
  \renewcommand{\arraystretch}{1.12}
  \begin{tabular*}{\linewidth}{@{}l@{\extracolsep{\fill}}rrrrrr@{}}
    \toprule
      & \multicolumn{2}{c}{Qwen3.5-27B (S1)}
      & \multicolumn{2}{c}{Qwen3.5-27B (S2)}
      & \multicolumn{2}{c}{Qwen3.8-27B} \\
    \cmidrule(lr){2-3}
    \cmidrule(lr){4-5}
    \cmidrule(lr){6-7}
    Environment
      & Instances & Rollout share (\%)
      & Instances & Rollout share (\%)
      & Instances & Rollout share (\%) \\
    \midrule
    Claw-Anything & 3,800 & 17.5 & 1,546 & 5.0 & 5,346 & 4.8 \\
    SWE       & 4,772 & 50.0 & 3,013 & 32.5 & 4,773 & 32.1 \\
    TMax          & 2,514 & 32.5 & 2,789 & 27.5 & 6,412 & 26.2 \\
    CodeWorld     & -- & -- & 3,104 & 35.0 & 4,344 & 36.9 \\
    \midrule
    Total & 11,086 & 100.0 & 10,452 & 100.0 & 20,875 & 100.0 \\
    \bottomrule
  \end{tabular*}
\end{table}

\subsection{Main Results}
\label{sec:main-results}

Table~\ref{tab:main-results} compares \name{} with Qwen baselines and frontier models across five benchmarks. With RL-only post-training, \name{} raises the Overall score of Qwen3.5-27B from 43.6 to 54.2 in S1. Expanding the existing task pools and introducing feature development in S2 further increases Overall to 60.1, surpassing the 49.8 achieved by Qwen3.6-27B. These results support scaling agent capabilities through incremental expansion of task instances and task types. Applied to Qwen3.8-27B, \name{} raises Overall from 61.4 to 67.0, improving performance on all five benchmarks and achieving the best overall performance among the evaluated 27B models and 
outperforms several substantially larger models (Figure~\ref{fig:cover-performance}(a)). \name{}-3.5 (S2) also achieves 64.6\% on Claw-Anything, exceeding all other models in Table~\ref{tab:main-results}.

Figure~\ref{fig:cover-performance}(b) compares mixed-task and single-task RL by cumulative policy updates, summing updates across three independent single-task runs. \name{} achieves higher mean Pass@1 across Claw-Eval, Terminal-Bench 2.1, and SWE-bench Verified with fewer cumulative policy updates.

\subsection{Ablation Studies}
\label{sec:ablations}

\begin{figure}[t]
  \centering
  \includegraphics[width=\linewidth]{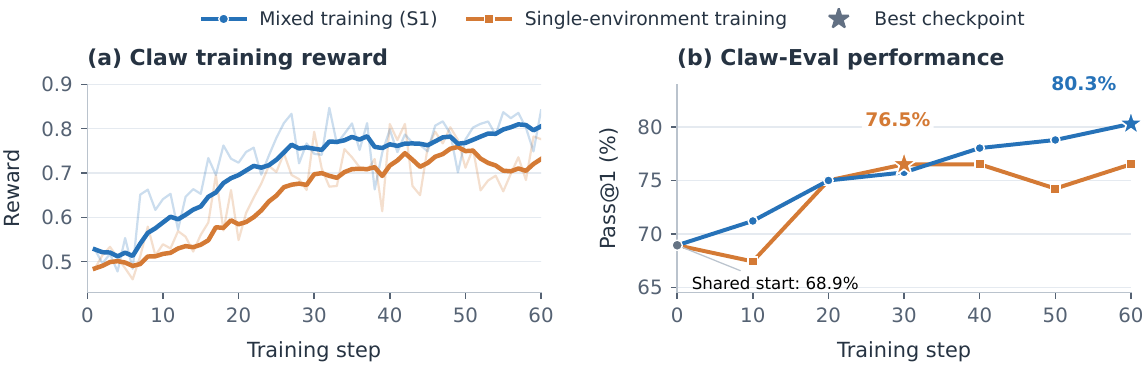}\par
  \vspace{-4pt}
  \caption{Target-task data efficiency of mixed-task RL. Mixed training achieves higher peak Claw-Eval Pass@1 than Claw-only training at matched cumulative Claw instance counts. (a) Training reward. (b) Claw-Eval Pass@1.}
  \label{fig:claw-mixed-training}
\end{figure}

\paragraph{Mixed-Task Training Improves Data Efficiency.} Figure~\ref{fig:cover-performance}(b) compares a single model jointly trained on Claw, SWE, and terminal tasks with three independently trained domain experts. Under the same total policy update budget, joint training achieves higher mean Pass@1, demonstrating greater training efficiency. We further examine target-task data efficiency using Claw as an example, comparing Claw-only training with mixed RL across the three stage 1 task types from the same base model. With both runs using the same cumulative number of Claw instances, mixed training yields faster reward improvement (Figure~\ref{fig:claw-mixed-training}). Peak Claw-Eval Pass@1 over the evaluated interval reaches 80.3\% for mixed training and 76.5\% for Claw-only training, a gain of 3.8 percentage points. These results suggest positive cross-task transfer and improved data efficiency on the target task.

\begin{figure}[t]
  \centering
  \includegraphics[width=\linewidth]{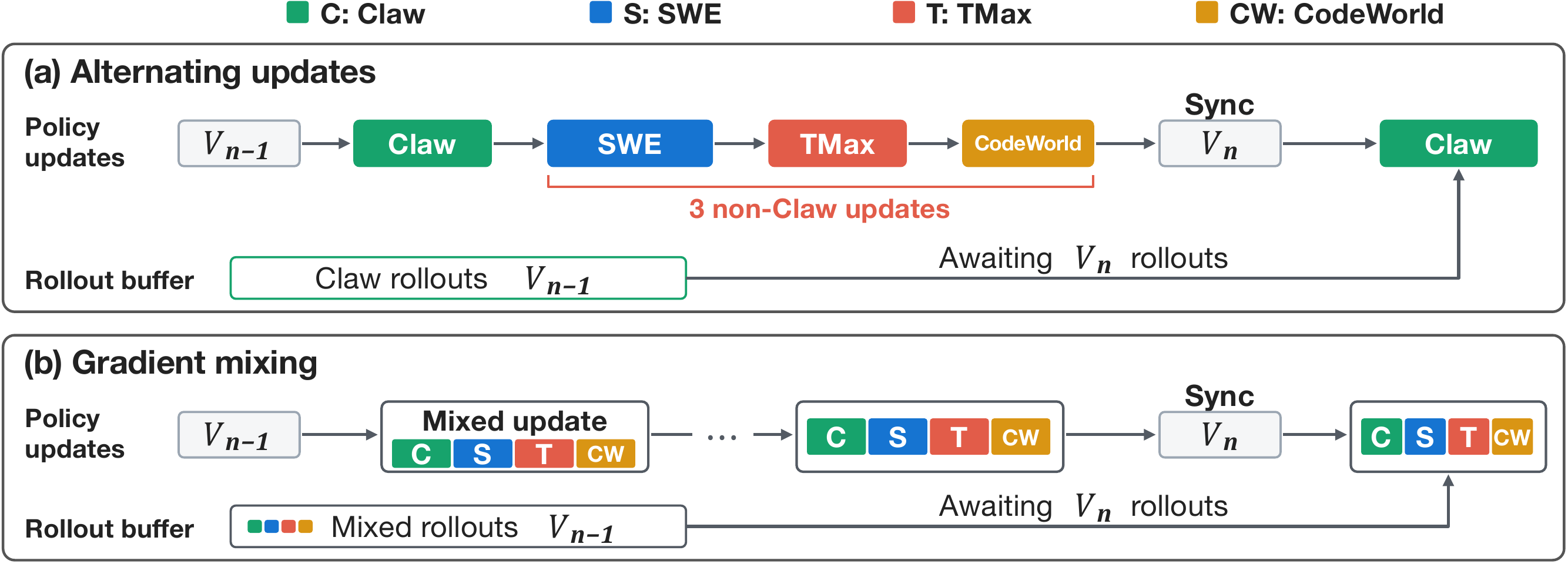}
  \caption{Policy lag under alternating updates and gradient mixing. $V_n$ denotes the $n$-th policy version synchronized to rollout workers; the version counter advances at each weight synchronization. (a) Alternating updates introduce three updates on other tasks between consecutive Claw updates. (b) Gradient mixing includes all task types in every update.}
  \label{fig:async-mixing-policy-lag}
\end{figure}
\begin{figure}[t]
  \centering
  \includegraphics[width=\linewidth]{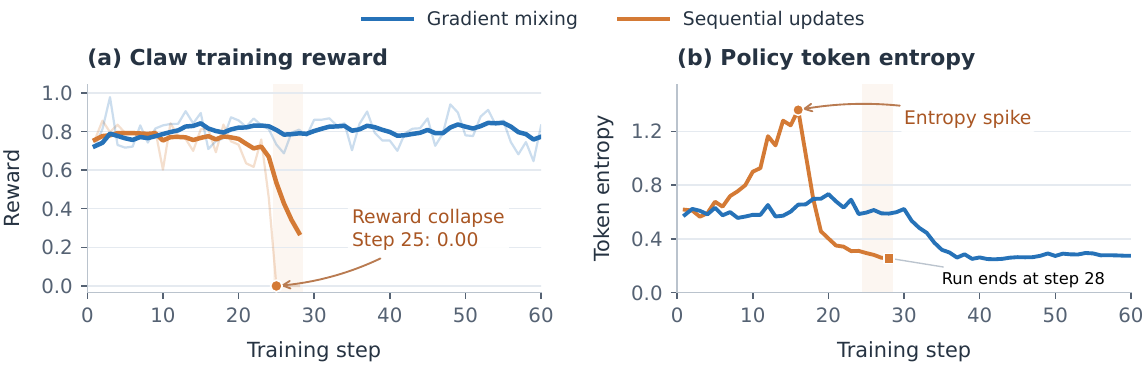}\par
  \caption{Training stability under alternating updates and gradient mixing in S2. Alternating updates lead to reward collapse, while gradient mixing maintains stable training. (a) Claw training reward. (b) Policy token entropy.}
  \label{fig:gradient-mixing-reward}
\end{figure}
\paragraph{Gradient Mixing Improves Training Stability.} We compare alternating updates and gradient mixing in stage 2 from the same final stage 1 checkpoint. Each update uses 32 instances with 8 rollouts each, totaling 256 trajectories. As shown in Figure~\ref{fig:async-mixing-policy-lag}, alternating updates use all 32 instances from one task type at a time, whereas gradient mixing uses 8 instances from each of the four task types in the same batch. We synchronize the rollout model with the training model after all four environments have been trained.

Under alternating updates, policy token entropy peaks at step 16 and then drops sharply; Claw reward falls to zero at step 25, accompanied by tool-use failures (Figure~\ref{fig:gradient-mixing-reward}). Gradient mixing maintains stable Claw reward as entropy decreases over training. These results support mixing learning signals across tasks within each update to improve training stability. 

This stability difference may arise from cross-task updates under asynchronous policy lag. Alternating updates cycle through Claw, SWE, terminal, and feature development tasks. Even after the rollout model is synchronized to $V_n$, the policy version published at synchronization $n$, the next Claw update may still rely on trajectories from $V_{n-1}$ while fresh rollouts are pending (Figure~\ref{fig:async-mixing-policy-lag}(a)). Between consecutive Claw updates, three updates driven solely by other tasks may alter tool-use and interaction patterns of the shared model. Consequently, even a one-version lag can correspond to substantial policy shift between the training and rollout models, potentially destabilizing the next Claw update.

With gradient mixing, each update jointly optimizes Claw and the other three task types (Figure~\ref{fig:async-mixing-policy-lag}(b)). Claw's learning signal may therefore counteract changes in tool-use and interaction patterns induced by other tasks. This may reduce the policy shift between the training model and the rollout model that generated the Claw trajectories, helping stabilize updates based on older rollouts.

\paragraph{Tri-Buffer Sampling Balances Learning Efficiency and Exploration.} We compare TBS with Vanilla sampling on feature development tasks, where 82.66\% of eligible instances have no successful rollout in initial screening. Uniform sampling therefore spends much of its rollout budget on initially unsolved instances, which can produce all-zero-reward groups and thus no reward-based policy-gradient signal. TBS instead assigns an effective 25\% Frontier quota when History is empty, prioritizing instances with occasional successes for reinforcing emerging capabilities, while reserving a fixed exploration quota for unsolved instances.

\begin{figure}[t]
  \centering
  \includegraphics[width=\linewidth]{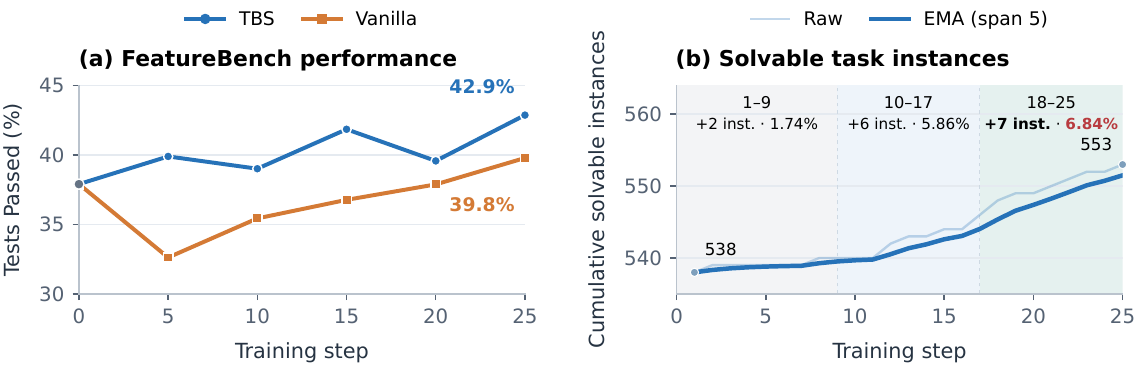}\par
  \caption{Learning efficiency and continued exploration with TBS on feature development tasks. (a) FeatureBench Tests Passed under TBS and Vanilla sampling. (b) Cumulative number of instances with at least one successful rollout under TBS. Interval labels show newly solved instance counts and their ratios to Frontier Buffer sampling attempts.}
  \label{fig:prd-sampling-ablation}
\end{figure}

Figure~\ref{fig:prd-sampling-ablation}(a) shows that TBS has a higher FeatureBench point estimate than Vanilla at every evaluated checkpoint after initialization. At step 25, TBS achieves 42.87\% Tests Passed, compared with 39.78\% for Vanilla, a gain of 3.09 percentage points with a task-paired 95\% confidence interval of $[-2.21,8.51]$. Meanwhile, Vanilla discovers successful rollouts for 33 initially unsolved instances from screening to step 25, compared with 15 for TBS, yet achieves lower FeatureBench performance. This difference is consistent with their sampling priorities: Vanilla allocates more opportunities to initially unsolved instances, whereas TBS reserves more capacity for reinforcing emerging capabilities. Vanilla therefore spend excessive rollout compute on all-zero-reward groups that provide no reward-based learning signal, reducing learning efficiency. TBS addresses this imbalance by explicitly limiting the exploration quota and reserving more capacity for reinforcement.

What's more, TBS continues to discover new solutions, with more discoveries in later training intervals (Figure~\ref{fig:prd-sampling-ablation}(b)). This pattern suggests that consolidating partially mastered instances may also help the policy solve previously unsolved ones, which then enter the Reinforcement Buffer for further training.

\subsection{Patterns of Reward Hacking}
\label{sec:reward-hacking}

During training and evaluation, models exploited exposed reference implementations to obtain rewards. Examples include retrieving fixes from Git history with \texttt{git show} in SWE-bench Verified and recovering target code from \texttt{build/lib}, source-code copies, wheel archives, or Conda caches in CodeWorld and FeatureBench. In one early mixed-training run, this behavior occurred in 12.33\% of all CodeWorld trajectories and 42.84\% of positive-reward trajectories. The identified leaks were removed, and test files were protected against tampering. Appendix~\ref{app:reward-hacking} presents representative cases and the corresponding environment safeguards.

Beyond the reward hacking, permissive LLM-based grading under-penalized invalid tool use, contributing to recurrent Claw-Anything reward collapse in S1. We strengthened execution checks, shortened rollout budgets, and enforced immediate termination with zero reward for malformed calls. After these repairs, training remained stable from step 170 to 270 (Appendix~\ref{app:claw-protocol-collapse}).

\section{Conclusion}
We present \name{}, a framework for scaling agent capabilities through direct MixRL with incremental expansion of task instances and task types. It combines asynchronous rollout generation and gradient mixing for efficient, stable joint training, while TBS balances exploration, reinforcement, and capability retention. With RL-only post-training, Qwen3.5-27B improves further from S1 to S2 as task instances and task types expand, surpassing Qwen3.6-27B in overall performance. Post-training Qwen3.8-27B yields the highest overall score among the evaluated 27B models, surpassing several substantially larger models. Joint training also outperforms independent domain specialists with fewer cumulative policy updates. Ablations show that joint training improves target-task data efficiency, gradient mixing stabilizes training, and TBS improves learning efficiency while sustaining exploration. These results motivate further scaling of task diversity, data volume, and compute within direct MixRL to advance general agent capabilities.

\clearpage

\bibliography{iclr2027_conference}

\begin{thebibliography}{44}
\providecommand{\natexlab}[1]{#1}
\providecommand{\url}[1]{\texttt{#1}}
\expandafter\ifx\csname urlstyle\endcsname\relax
  \providecommand{\doi}[1]{doi: #1}\else
  \providecommand{\doi}{doi: \begingroup \urlstyle{rm}\Url}\fi

\bibitem[Ahmadian et~al.(2024)Ahmadian, Cremer, Gall{\'e}, Fadaee, Kreutzer, Pietquin, et~al.]{ahmadian2024rloo}
Arash Ahmadian, Chris Cremer, Matthias Gall{\'e}, Marzieh Fadaee, Julia Kreutzer, Olivier Pietquin, et~al.
\newblock Back to basics: Revisiting {REINFORCE}-style optimization for learning from human feedback in {LLM}s.
\newblock In \emph{Proceedings of the 62nd Annual Meeting of the Association for Computational Linguistics (Volume 1: Long Papers)}, pp.\  12248--12267, Bangkok, Thailand, August 2024. Association for Computational Linguistics.
\newblock \doi{10.18653/v1/2024.acl-long.662}.
\newblock URL \url{https://aclanthology.org/2024.acl-long.662/}.

\bibitem[{Artificial Analysis}(2026)]{artificialanalysis2026terminalbench21}
{Artificial Analysis}.
\newblock {Terminal-Bench 2.1 Benchmark Leaderboard}.
\newblock Online leaderboard, 2026.
\newblock URL \url{https://artificialanalysis.ai/evaluations/terminalbench-2-1}.
\newblock Accessed September 26, 2026.

\bibitem[Bai et~al.(2024)Bai, Zhou, Cemri, Pan, Suhr, Levine, et~al.]{bai2024digirl}
Hao Bai, Yifei Zhou, Mert Cemri, Jiayi Pan, Alane Suhr, Sergey Levine, et~al.
\newblock {DigiRL}: Training in-the-wild device-control agents with autonomous reinforcement learning.
\newblock In \emph{Advances in Neural Information Processing Systems}, volume~37, pp.\  12461--12495. Curran Associates, Inc., 2024.
\newblock \doi{10.52202/079017-0397}.
\newblock URL \url{https://proceedings.neurips.cc/paper_files/paper/2024/hash/1704ddd0bb89f159dfe609b32c889995-Abstract-Conference.html}.

\bibitem[Cho et~al.(2024)Cho, Park, Lee, and Sung]{cho2024hard}
Myungsik Cho, Jongeui Park, Suyoung Lee, and Youngchul Sung.
\newblock Hard tasks first: Multi-task reinforcement learning through task scheduling.
\newblock In \emph{Proceedings of the 41st International Conference on Machine Learning}, volume 235 of \emph{Proceedings of Machine Learning Research}, pp.\  8556--8577. PMLR, 2024.
\newblock URL \url{https://proceedings.mlr.press/v235/cho24d.html}.

\bibitem[Chowdhury et~al.(2024)Chowdhury, Aung, Shern, Jaffe, Sherburn, Starace, et~al.]{chowdhury2024sweverified}
Neil Chowdhury, James Aung, Chan~Jun Shern, Oliver Jaffe, Dane Sherburn, Giulio Starace, et~al.
\newblock {Introducing SWE-bench Verified}.
\newblock OpenAI, August 2024.
\newblock URL \url{https://openai.com/index/introducing-swe-bench-verified/}.

\bibitem[{DeepSeek-AI}(2026)]{deepseek2026v4}
{DeepSeek-AI}.
\newblock {DeepSeek-V4}: Towards highly efficient million-token context intelligence.
\newblock \emph{arXiv preprint arXiv:2606.19348}, 2026.
\newblock \doi{10.48550/arXiv.2606.19348}.
\newblock URL \url{https://arxiv.org/abs/2606.19348}.

\bibitem[D'Eramo et~al.(2020)D'Eramo, Tateo, Bonarini, Restelli, and Peters]{deramo2020sharing}
Carlo D'Eramo, Davide Tateo, Andrea Bonarini, Marcello Restelli, and Jan Peters.
\newblock Sharing knowledge in multi-task deep reinforcement learning.
\newblock In \emph{International Conference on Learning Representations}, 2020.
\newblock URL \url{https://openreview.net/forum?id=rkgpv2VFvr}.

\bibitem[Espeholt et~al.(2018)Espeholt, Soyer, Munos, Simonyan, Mnih, Ward, Doron, Firoiu, Harley, Dunning, Legg, and Kavukcuoglu]{espeholt2018impala}
Lasse Espeholt, Hubert Soyer, Remi Munos, Karen Simonyan, Vlad Mnih, Tom Ward, Yotam Doron, Vlad Firoiu, Tim Harley, Iain Dunning, Shane Legg, and Koray Kavukcuoglu.
\newblock {IMPALA}: Scalable distributed deep-{RL} with importance weighted actor-learner architectures.
\newblock In \emph{Proceedings of the 35th International Conference on Machine Learning}, volume~80 of \emph{Proceedings of Machine Learning Research}, pp.\  1407--1416. PMLR, 2018.
\newblock URL \url{https://proceedings.mlr.press/v80/espeholt18a.html}.

\bibitem[Foster et~al.(2025)Foster, Sims, Forkel, and Foerster]{foster2025lilo}
Thomas Foster, Anya Sims, Johannes Forkel, and Jakob Foerster.
\newblock {LILO}: Learning to reason at the frontier of learnability.
\newblock In \emph{Advances in Neural Information Processing Systems}, volume~38, pp.\  48941--48974. Curran Associates, Inc., 2025.
\newblock \doi{10.52202/085713-1634}.
\newblock URL \url{https://proceedings.neurips.cc/paper_files/paper/2025/hash/45d5823950cab9a88e10c6e8dbff351c-Abstract-Conference.html}.

\bibitem[{GLM-5 Team}(2026)]{glm2026glm5}
{GLM-5 Team}.
\newblock {GLM-5: from Vibe Coding to Agentic Engineering}.
\newblock arXiv preprint arXiv:2602.15763, 2026.
\newblock URL \url{https://arxiv.org/abs/2602.15763}.

\bibitem[Hessel et~al.(2019)Hessel, Soyer, Espeholt, Czarnecki, Schmitt, and van Hasselt]{hessel2019popart}
Matteo Hessel, Hubert Soyer, Lasse Espeholt, Wojciech Czarnecki, Simon Schmitt, and Hado van Hasselt.
\newblock Multi-task deep reinforcement learning with {PopArt}.
\newblock In \emph{Proceedings of the AAAI Conference on Artificial Intelligence}, volume~33, pp.\  3796--3803, 2019.
\newblock \doi{10.1609/aaai.v33i01.33013796}.
\newblock URL \url{https://ojs.aaai.org/index.php/AAAI/article/view/4266}.

\bibitem[Huang et~al.(2026)Huang, Xue, Peng, Ding, Yang, Fan, et~al.]{huang2026evocua}
Mianqiu Huang, Taofeng Xue, Chong Peng, Jinrui Ding, Jie Yang, Sicheng Fan, et~al.
\newblock {EvoCUA-1.5: Online Reinforcement Learning for Multi-turn Computer-Use Agents}.
\newblock arXiv preprint arXiv:2607.09773, 2026.
\newblock URL \url{https://arxiv.org/abs/2607.09773}.

\bibitem[Ivison et~al.(2026)Ivison, Yin, Shao, Xiao, Lambert, and Hajishirzi]{ivison2026tmax}
Hamish Ivison, Junjie~Oscar Yin, Rulin Shao, Teng Xiao, Nathan Lambert, and Hannaneh Hajishirzi.
\newblock {Tmax: A simple recipe for terminal agents}.
\newblock arXiv preprint arXiv:2606.23321, 2026.
\newblock URL \url{https://arxiv.org/abs/2606.23321}.

\bibitem[Jimenez et~al.(2024)Jimenez, Yang, Wettig, Yao, Pei, Press, and Narasimhan]{jimenez2024swe}
Carlos~E Jimenez, John Yang, Alexander Wettig, Shunyu Yao, Kexin Pei, Ofir Press, and Karthik Narasimhan.
\newblock Swe-bench: Can language models resolve real-world github issues?
\newblock In \emph{International Conference on Learning Representations}, volume 2024, pp.\  54107--54157, 2024.

\bibitem[{Kimi Team}(2026)]{kimi2026k3}
{Kimi Team}.
\newblock {Kimi K3}: Open frontier intelligence.
\newblock \emph{arXiv preprint arXiv:2607.24653}, 2026.
\newblock \doi{10.48550/arXiv.2607.24653}.
\newblock URL \url{https://arxiv.org/abs/2607.24653}.

\bibitem[Kong et~al.(2026)Kong, Guo, Xi, Wang, Wang, Cai, et~al.]{kong2026cdas}
Deyang Kong, Qi~Guo, Xiangyu Xi, Wei Wang, Jingang Wang, Xunliang Cai, et~al.
\newblock {Rethinking the Sampling Criteria in Reinforcement Learning for LLM Reasoning: A Competence-Difficulty Alignment Perspective}.
\newblock In \emph{Proceedings of the AAAI Conference on Artificial Intelligence}, volume~40, pp.\  31438--31446, 2026.
\newblock \doi{10.1609/aaai.v40i37.40408}.
\newblock URL \url{https://ojs.aaai.org/index.php/AAAI/article/view/40408}.

\bibitem[Lai et~al.(2026)Lai, Liu, Zhao, Xu, Zhang, Jing, et~al.]{lai2026computerrl}
Hanyu Lai, Xiao Liu, Yanxiao Zhao, Han Xu, Hanchen Zhang, Bohao Jing, et~al.
\newblock {ComputerRL: Scaling End-to-End Online Reinforcement Learning for Computer Use Agents}.
\newblock In \emph{International Conference on Learning Representations}, pp.\  23553--23591, 2026.
\newblock URL \url{https://proceedings.iclr.cc/paper_files/paper/2026/hash/28707388472adc117714e83d44bfaebb-Abstract-Conference.html}.

\bibitem[Li et~al.(2025)Li, Hu, Shang, Wu, Liu, Liu, et~al.]{li2025dart}
Pengxiang Li, Zechen Hu, Zirui Shang, Jingrong Wu, Yang Liu, Hui Liu, et~al.
\newblock {Efficient Multi-turn RL for GUI Agents via Decoupled Training and Adaptive Data Curation}.
\newblock arXiv preprint arXiv:2509.23866, 2025.
\newblock URL \url{https://arxiv.org/abs/2509.23866}.

\bibitem[Lin et~al.(2026)Lin, Liang, Wang, Gu, Cheng, Chen, et~al.]{lin2026clawanything}
Yusong Lin, Xinyuan Liang, Haiyang Wang, Qipeng Gu, Siqi Cheng, Jiangui Chen, et~al.
\newblock {Claw-Anything: Benchmarking Always-On Personal Assistants with Broader Access to User's Digital World}.
\newblock arXiv preprint arXiv:2605.26086, 2026.
\newblock URL \url{https://arxiv.org/abs/2605.26086}.

\bibitem[Luo et~al.(2025)Luo, Jain, Singh, Tan, Patel, Wu, Ariyak, Cai, Venkat, Zhu, Athiwaratkun, Roongta, Zhang, Li, Popa, Sen, and Stoica]{luo2025deepswe}
Michael Luo, Naman Jain, Jaskirat Singh, Sijun Tan, Ameen Patel, Qingyang Wu, Alpay Ariyak, Colin Cai, Tarun Venkat, Shang Zhu, Ben Athiwaratkun, Manan Roongta, Ce~Zhang, Li~Erran Li, Raluca~Ada Popa, Koushik Sen, and Ion Stoica.
\newblock {DeepSWE}: Training a fully open-sourced, state-of-the-art coding agent by scaling {RL}.
\newblock Together AI Blog, July 2025.
\newblock URL \url{https://www.together.ai/blog/deepswe}.

\bibitem[Lv et~al.(2026)Lv, Liu, Ren, Lai, Jing, Zhang, et~al.]{lv2026scalecua}
Bowen Lv, Xiao Liu, Yanyu Ren, Hanyu Lai, Bohao Jing, Hanchen Zhang, et~al.
\newblock {SCALECUA: Scaling Computer Use Agents with Verifiable Task Synthesis and Efficient Online RL}.
\newblock arXiv preprint arXiv:2607.11185, 2026.
\newblock URL \url{https://arxiv.org/abs/2607.11185}.

\bibitem[Merrill et~al.(2026)Merrill, Shaw, Carlini, Li, Raj, Bercovich, et~al.]{merrill2026terminalbench}
Mike Merrill, Alexander Shaw, Nicholas Carlini, Boxuan Li, Harsh Raj, Ivan Bercovich, et~al.
\newblock {Terminal-Bench: Benchmarking Agents on Hard, Realistic Tasks in Command Line Interfaces}.
\newblock In \emph{International Conference on Learning Representations}, pp.\  40903--40986, 2026.
\newblock URL \url{https://proceedings.iclr.cc/paper_files/paper/2026/hash/444a3737adaee10d86ad2ef5f74468e6-Abstract-Conference.html}.

\bibitem[{MiMo Team}(2026)]{xiaomi2026mimov2flash}
{MiMo Team}.
\newblock {MiMo-V2-Flash} technical report.
\newblock \emph{arXiv preprint arXiv:2601.02780}, 2026.
\newblock URL \url{https://arxiv.org/abs/2601.02780}.

\bibitem[Qi et~al.(2025)Qi, Liu, Iong, Lai, Sun, Sun, et~al.]{qi2025webrl}
Zehan Qi, Xiao Liu, Iat~Long Iong, Hanyu Lai, Xueqiao Sun, Jiadai Sun, et~al.
\newblock {WebRL: Training LLM Web Agents via Self-Evolving Online Curriculum Reinforcement Learning}.
\newblock In \emph{International Conference on Learning Representations}, pp.\  79791--79821, 2025.
\newblock URL \url{https://proceedings.iclr.cc/paper_files/paper/2025/hash/c66e1fcc9691aae706250638f36f681b-Abstract-Conference.html}.

\bibitem[{Qwen Team}(2026)]{qwen2026qwen35}
{Qwen Team}.
\newblock {Qwen3.5: Towards Native Multimodal Agents}, February 2026.
\newblock URL \url{https://qwen.ai/blog?id=qwen3.5}.

\bibitem[Schaul et~al.(2016)Schaul, Quan, Antonoglou, and Silver]{schaul2016prioritized}
Tom Schaul, John Quan, Ioannis Antonoglou, and David Silver.
\newblock Prioritized experience replay.
\newblock In \emph{International Conference on Learning Representations}, 2016.
\newblock URL \url{https://arxiv.org/abs/1511.05952}.

\bibitem[Sheng et~al.(2025)Sheng, Zhang, Ye, Wu, Zhang, Zhang, et~al.]{sheng2025hybridflow}
Guangming Sheng, Chi Zhang, Zilingfeng Ye, Xibin Wu, Wang Zhang, Ru~Zhang, et~al.
\newblock {HybridFlow: A Flexible and Efficient RLHF Framework}.
\newblock In \emph{Proceedings of the Twentieth European Conference on Computer Systems}, pp.\  1279--1297, 2025.
\newblock \doi{10.1145/3689031.3696075}.
\newblock URL \url{https://doi.org/10.1145/3689031.3696075}.

\bibitem[Song et~al.(2026)Song, Bai, Yang, Li, Deng, He, Zhang, Cheng, Xing, Yun, et~al.]{song2026clawgym}
Huatong Song, Fei Bai, Ming Yang, Renyuan Li, Jia Deng, Jujie He, Zhange Zhang, Daixuan Cheng, Yan Xing, Qi~Yun, et~al.
\newblock Clawgym ii: Exploring black-box rl on agent harness.
\newblock \emph{arXiv preprint arXiv:2608.16798}, 2026.

\bibitem[Teh et~al.(2017)Teh, Bapst, Czarnecki, Quan, Kirkpatrick, Hadsell, Heess, and Pascanu]{teh2017distral}
Yee~Whye Teh, Victor Bapst, Wojciech~Marian Czarnecki, John Quan, James Kirkpatrick, Raia Hadsell, Nicolas Heess, and Razvan Pascanu.
\newblock {Distral}: Robust multitask reinforcement learning.
\newblock In \emph{Advances in Neural Information Processing Systems}, volume~30, 2017.
\newblock URL \url{https://arxiv.org/abs/1707.04175}.

\bibitem[{Terminal-Bench Team}(2026)]{terminalbench2026v21}
{Terminal-Bench Team}.
\newblock {Terminal-Bench 2.1}, 2026.
\newblock URL \url{https://www.tbench.ai/news/terminal-bench-2-1}.

\bibitem[{The slime Team}(2025)]{slimeteam2025slime}
{The slime Team}.
\newblock {slime: An SGLang-Native Post-Training Framework for RL Scaling}.
\newblock LMSYS Org blog, July 2025.
\newblock URL \url{https://www.lmsys.org/blog/2025-07-09-slime/}.

\bibitem[{Vals AI}(2026)]{valsai2026swebench}
{Vals AI}.
\newblock {SWE-bench Verified}.
\newblock Online leaderboard, 2026.
\newblock URL \url{https://www.vals.ai/benchmarks/swebench}.
\newblock Accessed September 26, 2026.

\bibitem[Vithayathil~Varghese \& Mahmoud(2020)Vithayathil~Varghese and Mahmoud]{varghese2020survey}
Nelson Vithayathil~Varghese and Qusay~H. Mahmoud.
\newblock A survey of multi-task deep reinforcement learning.
\newblock \emph{Electronics}, 9\penalty0 (9):\penalty0 1363, 2020.
\newblock \doi{10.3390/electronics9091363}.
\newblock URL \url{https://www.mdpi.com/2079-9292/9/9/1363}.

\bibitem[Wang et~al.(2025)Wang, Wang, Wang, Zhang, Li, Yang, et~al.]{wang2025ragen}
Zihan Wang, Kangrui Wang, Qineng Wang, Pingyue Zhang, Linjie Li, Zhengyuan Yang, et~al.
\newblock {RAGEN: Understanding Self-Evolution in LLM Agents via Multi-Turn Reinforcement Learning}.
\newblock arXiv preprint arXiv:2504.20073, 2025.
\newblock URL \url{https://arxiv.org/abs/2504.20073}.

\bibitem[Xi et~al.(2025)Xi, Ding, Chen, Hong, Guo, Wang, et~al.]{xi2025agentgym}
Zhiheng Xi, Yiwen Ding, Wenxiang Chen, Boyang Hong, Honglin Guo, Junzhe Wang, et~al.
\newblock {A}gent{G}ym: Evaluating and training large language model-based agents across diverse environments.
\newblock In \emph{Proceedings of the 63rd Annual Meeting of the Association for Computational Linguistics (Volume 1: Long Papers)}, pp.\  27914--27961, Vienna, Austria, July 2025. Association for Computational Linguistics.
\newblock ISBN 979-8-89176-251-0.
\newblock \doi{10.18653/v1/2025.acl-long.1355}.
\newblock URL \url{https://aclanthology.org/2025.acl-long.1355/}.

\bibitem[Xi et~al.(2026)Xi, Guo, Liu, Zhang, Fan, Zhang, et~al.]{xi2026generalization}
Zhiheng Xi, Xin Guo, Jiaqi Liu, Jiazheng Zhang, Yutao Fan, Zhihao Zhang, et~al.
\newblock {Can RL Improve Generalization of LLM Agents? An Empirical Study}.
\newblock arXiv preprint arXiv:2603.12011, 2026.
\newblock URL \url{https://arxiv.org/abs/2603.12011}.

\bibitem[Xue et~al.(2026)Xue, Liao, Shi, Wang, Zhang, Song, et~al.]{xue2026acurl}
Tianci Xue, Zeyi Liao, Tianneng Shi, Zilu Wang, Kai Zhang, Dawn Song, et~al.
\newblock {Autonomous Continual Learning for Environment Adaptation of Computer-Use Agents}.
\newblock arXiv preprint arXiv:2602.10356, 2026.
\newblock URL \url{https://arxiv.org/abs/2602.10356}.

\bibitem[Yang et~al.(2024)Yang, Jimenez, Wettig, Lieret, Yao, Narasimhan, et~al.]{yang2024sweagent}
John Yang, Carlos Jimenez, Alexander Wettig, Kilian Lieret, Shunyu Yao, Karthik Narasimhan, et~al.
\newblock {SWE-agent}: Agent-computer interfaces enable automated software engineering.
\newblock In \emph{Advances in Neural Information Processing Systems}, volume~37, pp.\  50528--50652. Curran Associates, Inc., 2024.
\newblock \doi{10.52202/079017-1601}.
\newblock URL \url{https://proceedings.neurips.cc/paper_files/paper/2024/hash/5a7c947568c1b1328ccc5230172e1e7c-Abstract-Conference.html}.

\bibitem[Ye et~al.(2026)Ye, Li, Yang, Liu, Yao, Lv, et~al.]{ye2026claweval}
Bowen Ye, Rang Li, Qibin Yang, Yuanxin Liu, Linli Yao, Hanglong Lv, et~al.
\newblock {Claw-Eval: Towards Trustworthy Evaluation of Autonomous Agents}.
\newblock arXiv preprint arXiv:2604.06132, 2026.
\newblock URL \url{https://arxiv.org/abs/2604.06132}.

\bibitem[Yu et~al.(2025)Yu, Zhang, Zhu, Yuan, Zuo, Yue, et~al.]{yu2025dapo}
Qiying Yu, Zheng Zhang, Ruofei Zhu, Yufeng Yuan, Xiaochen Zuo, Yu~Yue, et~al.
\newblock {DAPO: An Open-Source LLM Reinforcement Learning System at Scale}.
\newblock In \emph{Advances in Neural Information Processing Systems}, volume~38, pp.\  113222--113244. Curran Associates, Inc., 2025.
\newblock \doi{10.52202/085713-3775}.
\newblock URL \url{https://proceedings.neurips.cc/paper_files/paper/2025/hash/a4277440d50f1f15d2cb4c14f7e0c0d2-Abstract-Conference.html}.

\bibitem[Yu et~al.(2020)Yu, Kumar, Gupta, Levine, Hausman, and Finn]{yu2020gradient}
Tianhe Yu, Saurabh Kumar, Abhishek Gupta, Sergey Levine, Karol Hausman, and Chelsea Finn.
\newblock {Gradient Surgery for Multi-Task Learning}.
\newblock In \emph{Advances in Neural Information Processing Systems}, volume~33, pp.\  5824--5836, 2020.
\newblock URL \url{https://proceedings.neurips.cc/paper/2020/hash/3fe78a8acf5fda99de95303940a2420c-Abstract.html}.

\bibitem[Zeng et~al.(2024)Zeng, Liu, Lu, Wang, Liu, Dong, et~al.]{zeng2024agenttuning}
Aohan Zeng, Mingdao Liu, Rui Lu, Bowen Wang, Xiao Liu, Yuxiao Dong, et~al.
\newblock {A}gent{T}uning: Enabling generalized agent abilities for {LLM}s.
\newblock In \emph{Findings of the Association for Computational Linguistics: ACL 2024}, pp.\  3053--3077, Bangkok, Thailand, August 2024. Association for Computational Linguistics.
\newblock \doi{10.18653/v1/2024.findings-acl.181}.
\newblock URL \url{https://aclanthology.org/2024.findings-acl.181/}.

\bibitem[Zhang et~al.(2025)Zhang, Liu, Lv, Sun, Jing, Iong, et~al.]{zhang2025agentrl}
Hanchen Zhang, Xiao Liu, Bowen Lv, Xueqiao Sun, Bohao Jing, Iat~Long Iong, et~al.
\newblock {AgentRL: Scaling Agentic Reinforcement Learning with a Multi-Turn, Multi-Task Framework}.
\newblock arXiv preprint arXiv:2510.04206, 2025.
\newblock URL \url{https://arxiv.org/abs/2510.04206}.

\bibitem[Zhou et~al.(2026)Zhou, Zhang, Wang, Hao, Wang, Han, et~al.]{zhou2026featurebench}
Qixing Zhou, Jiacheng Zhang, Haiyang Wang, Rui Hao, Jiahe Wang, Minghao Han, et~al.
\newblock {FeatureBench: Benchmarking Agentic Coding for Complex Feature Development}.
\newblock In \emph{International Conference on Learning Representations}, pp.\  22019--22055, 2026.
\newblock URL \url{https://proceedings.iclr.cc/paper_files/paper/2026/hash/25203d1cc8c58381eab578f4fcf9c4f8-Abstract-Conference.html}.

\end{thebibliography}
\bibliographystyle{iclr2027_conference}

\clearpage
\appendix

\section{Environment and Dataset Details}
\label{app:environments}
\subsection{Dataset Composition and Preprocessing}
Qwen3.5-27B training proceeds in two stages, with additional task instances and the CodeWorld environment introduced in stage 2. Table~\ref{tab:app-dataset-composition} summarizes screening outcomes for the instances added at each stage and the cumulative eligible task pools. Each instance undergoes eight independent rollouts and is classified as Unsolved (0 successes), Partial (1--7), or Solved (8). Instances that cannot complete screening because of execution errors are classified as Invalid. TMax excludes Solved instances to reduce repeated training on tasks the model can already solve reliably, whereas SWE and CodeWorld retain all valid instances across the three success-rate categories. Claw-Anything uses uniform sampling without success-rate screening, so these categories are not reported for it in the corresponding table.

\begin{table}[htbp]
  \centering
  \caption{New task instances, screening outcomes, and cumulative eligible instance pools.}
  \label{tab:app-dataset-composition}
  \small
  \setlength{\tabcolsep}{4pt}
  \renewcommand{\arraystretch}{1.15}
  \resizebox{\linewidth}{!}{%
  \begin{tabular}{@{}llrrrrrrr@{}}
    \toprule
    Environment & Stage & New instances & \shortstack{Unsolved\\0/8} & \shortstack{Partial\\1--7/8} & \shortstack{Solved\\8/8} & Invalid & Retained & \shortstack{Cumulative\\pool} \\
    \midrule
    Claw-Anything & S1 & 3,800 & -- & -- & -- & -- & 3,800 & 3,800 \\
     & S2 & 1,546 & -- & -- & -- & -- & 1,546 & 5,346 \\
    \midrule
    SWE & S1 & 4,886 & 3,157 & 1,266 & 349 & 114 & 4,772 & 4,772 \\
     & S2 & 3,020 & 1,551 & 428 & 1,034 & 7 & 3,013 & 7,785 \\
    \midrule
    TMax & S1 & 5,370 & 1,081 & 1,433 & 2,856 & 0 & 2,514 & 2,514 \\
     & S2 & 6,032 & 1,229 & 1,560 & 3,243 & 0 & 2,789 & 5,303 \\
    \midrule
    CodeWorld & S2 & 3,122 & 2,566 & 464 & 74 & 18 & 3,104 & 3,104 \\
    \bottomrule
  \end{tabular}%
  }
\end{table}

\subsection{Screening Configuration}
To initialize sampling weights for TBS, each task instance is evaluated with eight independent rollouts to estimate its initial success rate. TMax uses Qwen3.5-27B, while SWE and CodeWorld use the faster Qwen3.6-35B-A3B to reduce screening costs. Table~\ref{tab:app-screening} summarizes the inference settings. P/R denotes the token budgets for the initial task prompt and cumulative model responses, respectively, with $1\mathrm{K}=1024$ tokens. $T$, $p$, and $k$ denote temperature, top-$p$, and top-$k$; $k=0$ disables top-$k$ truncation. All three environments use an 8K-token limit per model call and a 300-second tool timeout. TMax uses the JSON-based Terminus-2 interface with automatic summarization disabled, while SWE and CodeWorld use SWE-Agent.

\begin{table}[htbp]
  \centering
  \caption{Task instance screening configurations; eight rollouts per instance.}
  \label{tab:app-screening}
  \small
  \setlength{\tabcolsep}{4pt}
  \renewcommand{\arraystretch}{1.15}
  \resizebox{\linewidth}{!}{%
  \begin{tabular}{@{}llllrrl@{}}
    \toprule
    Environment & Screening model & Agent & P/R (K) & Max steps & Timeout (s) & $T / p / k$ \\
    \midrule
    TMax & Qwen3.5-27B & Terminus-2 & 16/64 & 100 & 5,400 & 1.0/1.0/0 \\
    SWE & Qwen3.6-35B-A3B & SWE-Agent & 9/128 & 150 & 5,400 & 1.0/0.95/20 \\
    CodeWorld & Qwen3.6-35B-A3B & SWE-Agent & 16/128 & 150 & 5,400 & 1.0/0.95/20 \\
    \bottomrule
  \end{tabular}%
  }
\end{table}

\subsection{Task Interfaces and Rewards}
\begin{figure}[p]
  \centering
  \includegraphics[width=\linewidth]{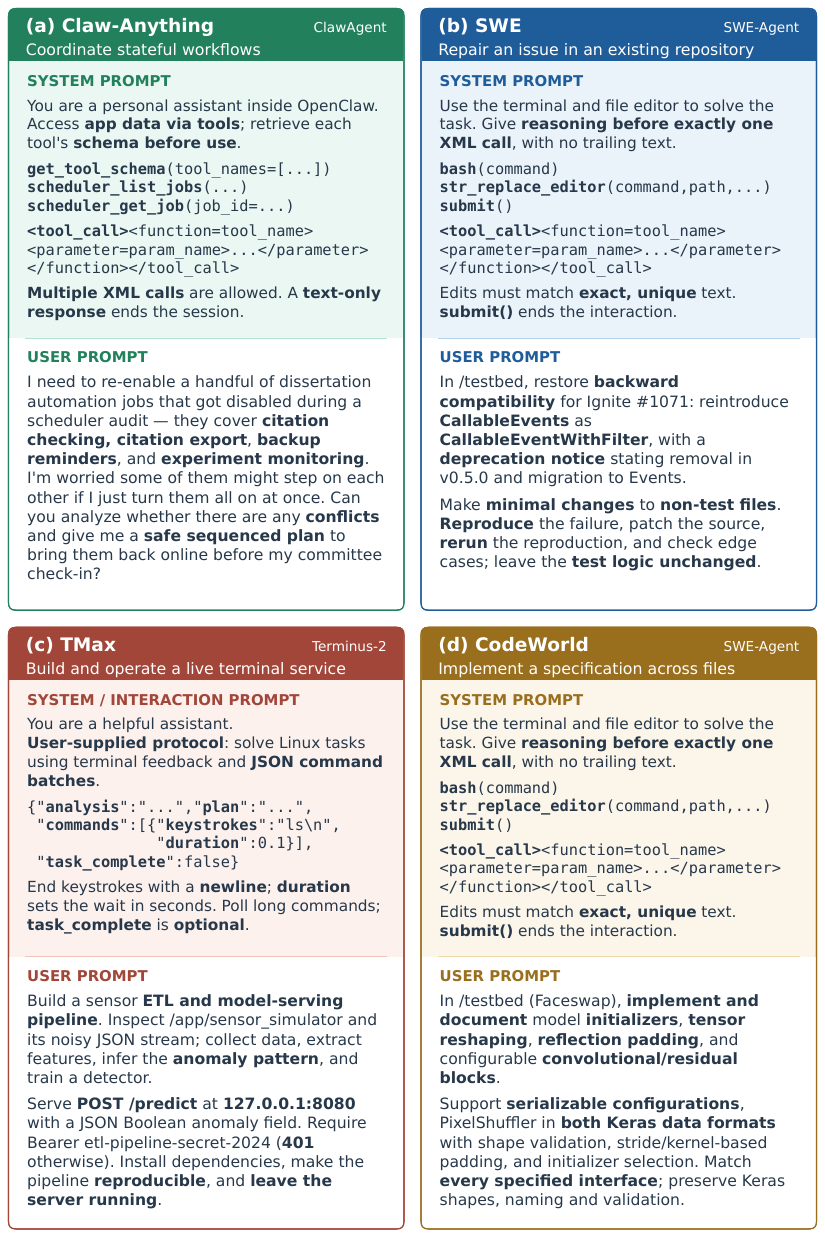}
  \caption{Condensed system and user prompts from four training environments, with illustrative tool interfaces. TMax supplies its interaction protocol in the user message.}
  \label{fig:app-environment-prompts}
\end{figure}

The four environments differ in task objectives, tool interfaces, and reward functions. Figure~\ref{fig:app-environment-prompts} presents excerpts of their system and task prompts from actual stage 2 training trajectories.

\paragraph{Claw-Anything.}
This environment covers office tasks involving email, calendars, documents, and automated workflows. The agent retrieves the required tool schemas, queries or modifies environment state, and produces a final response according to the task requirements. Rewards combine task graders, environment state, and LLM judgments, using Qwen3.5-397B-A17B as the judge during training and Kimi-K2.6 for Claw-Anything evaluation. For a task score $s\in[0,1]$, the base reward is 1 if $s\geq0.70$ and $s$ otherwise, providing finer-grained feedback for unsuccessful attempts.

\paragraph{SWE.}
These repository-level issue-resolution tasks require the agent to modify and debug source code based on an issue description, using the command execution, file editing, and submission tools provided by SWE-Agent. The task's verification script assigns a reward of 1 if verification passes and 0 otherwise. Protected test files are restored before verification to prevent changes to test logic from affecting the result.

\paragraph{TMax.}
This environment provides Linux terminal tasks involving installation and configuration, data processing, service deployment, and program debugging. Through Terminus-2, the agent produces analysis, execution plans, and batches of shell commands, then continues based on terminal feedback. At task termination, the built-in verification script returns a reward in $[0,1]$. During screening, a reward of 1 counts as a successful rollout.

\paragraph{CodeWorld.}
This environment requires the agent to implement software changes across multiple files from product requirements documents, using SWE-Agent for interaction. Verification applies the candidate patch in an isolated environment and runs fail-to-pass (F2P) tests for new functionality and pass-to-pass (P2P) tests for existing functionality. The reward is 1 only if all required tests pass without verification errors, and 0 otherwise. This task-level, rule-based reward requires both implementing the new functionality and preserving existing behavior.

\section{Training and Implementation Details}
\label{app:training}
\subsection{Hyperparameters and Resource Allocation}
\label{app:optimization-budgets}

\subsubsection{Optimization and Execution Resources}
Training uses 64 accelerators with tensor and pipeline parallelism both set to 8 ($\mathrm{TP}=\mathrm{PP}=8$) and context parallelism set to 1. Inference replicas use $\mathrm{TP}=4$ and $\mathrm{PP}=1$ on separate devices. The inference resource allocation and per-environment concurrency can be adjusted to available resources, independently of the per-environment training batch sizes.

Table~\ref{tab:app-runtime-resources} summarizes resource allocations for both training stages. Each slot accommodates one task instance with eight concurrent rollouts. Slot counts are configurable and specify concurrent task groups, rather than dedicated devices or instances per policy update. In S1, inference expands from 32 to 40 devices after 20 weight synchronization cycles. Slots are redistributed after cycle 170 to accommodate changes in per-environment rollout speed as response lengths evolve during training. S2 uses all four environments. The overall task pool sizes and the corresponding rollout budgets are reported in Tables~\ref{tab:app-dataset-composition} and~\ref{tab:app-training-rollout}, respectively.

\begin{table}[htbp]
  \centering
  \caption{Execution resources and concurrent task-group slots. Slot order: Claw / SWE / TMax / CodeWorld.}
  \label{tab:app-runtime-resources}
  \small
  \setlength{\tabcolsep}{4pt}
  \renewcommand{\arraystretch}{1.12}
  \begin{tabular*}{\linewidth}{@{\extracolsep{\fill}}lrrcr@{}}
    \toprule
    Run segment & Training devices & Inference devices & Slots & Total groups \\
    \midrule
    S1 (initial) & 64 & 32 & 4 / 16 / 12 / -- & 32 \\
    S1 (resumed) & 64 & 40 & 7 / 20 / 13 / -- & 40 \\
    S1 (late) & 64 & 40 & 4 / 19 / 17 / -- & 40 \\
    S2 & 64 & 32 & 2 / 13 / 11 / 14 & 40 \\
    \bottomrule
  \end{tabular*}
\end{table}

Policy updates use RLOO for advantage estimation, with eight rollouts per task instance and 256 trajectories from 32 instances per update. Table~\ref{tab:app-optimization} summarizes the hyperparameters. The policy loss is summed over valid response tokens within each trajectory and then averaged across trajectories:
\begin{equation}
  \mathcal{L}_{\mathrm{train}}
  = \frac{1}{256}\sum_{i=1}^{256}\sum_{t\in\mathcal{T}_i}\ell_{i,t},
  \label{eq:trajectory-loss-aggregation}
\end{equation}
where $\mathcal{T}_i$ contains the valid response tokens contributing to the loss in trajectory $i$, and $\ell_{i,t}$ is the corresponding token-level policy loss. KL loss and entropy regularization are disabled, as is thinking mode during rollouts.

Training uses BF16 computation and Adam with a constant learning rate and no warmup. Gradient checkpointing, optimizer-state CPU offloading, and output-loss computation in chunks of 1,024 tokens support long trajectories. The model's native multi-token prediction (MTP) auxiliary loss is enabled with a coefficient of 0.1, with gradients stopped at its encoder inputs. Equation~(\ref{eq:trajectory-loss-aggregation}) describes the aggregation of the policy loss only. MTP speculative decoding is disabled during inference.

\begin{table}[htbp]
  \centering
  \caption{Optimization and synchronization settings.}
  \label{tab:app-optimization}
  \small
  \setlength{\tabcolsep}{6pt}
  \renewcommand{\arraystretch}{1.15}
  \begin{tabular*}{\linewidth}{@{\extracolsep{\fill}}lcc@{}}
    \toprule
    Parameter & S1 & S2 \\
    \midrule
    Optimizer & \multicolumn{2}{c}{Adam} \\
    Learning rate & \multicolumn{2}{c}{$10^{-6}$} \\
    Learning-rate schedule / warmup & \multicolumn{2}{c}{Constant / None} \\
    Adam $\beta_1,\beta_2$ & \multicolumn{2}{c}{0.9, 0.999} \\
    Weight decay & \multicolumn{2}{c}{0.01} \\
    Gradient norm clipping & \multicolumn{2}{c}{1.0} \\
    Compute precision & \multicolumn{2}{c}{BF16} \\
    Micro-batch size per device & \multicolumn{2}{c}{1 trajectory} \\
    Lower clipping parameter & \multicolumn{2}{c}{0.20} \\
    Upper clipping parameter & \multicolumn{2}{c}{0.28} \\
    Advantage estimator & \multicolumn{2}{c}{RLOO} \\
    Instances per update & \multicolumn{2}{c}{32} \\
    Rollouts per instance & \multicolumn{2}{c}{8} \\
    KL loss & \multicolumn{2}{c}{Disabled} \\
    Entropy coefficient & \multicolumn{2}{c}{0} \\
    Loss aggregation & \multicolumn{2}{c}{Token sum, then trajectory mean} \\
    MTP training / loss coefficient & \multicolumn{2}{c}{Enabled / 0.1} \\
    MTP speculative decoding & \multicolumn{2}{c}{Disabled} \\
    Thinking mode & \multicolumn{2}{c}{Disabled} \\
    Global prompt/response budget (K) & 16/128 & 16/96 \\
    Updates per weight synchronization & 3 & 2 \\
    \bottomrule
  \end{tabular*}
\end{table}

\subsubsection{Rollout Budgets}
Table~\ref{tab:app-training-rollout} summarizes environment-specific rollout budgets. Each policy update uses 32 task instances, with eight rollouts each. Global response budgets are 128K and 96K tokens, respectively. Claw's cumulative response budget is reduced from 128K to 64K at step 130 of S1 and retained in S2, which also limits Claw trajectories to 50 interaction steps. P/R denotes the prompt and cumulative response budgets, and Turn denotes the per-call generation limit. A dash indicates no separate per-call limit; the overall token budget still applies.

\begin{table}[htbp]
  \centering
  \caption{Environment-specific rollout budgets during training.}
  \label{tab:app-training-rollout}
  \small
  \setlength{\tabcolsep}{4pt}
  \renewcommand{\arraystretch}{1.15}
  \resizebox{\linewidth}{!}{%
  \begin{tabular}{@{}lllllllll@{}}
    \toprule
    Stage & Environment & Agent & P/R (K) & Turn (K) & Max steps & \shortstack{Trajectory\\timeout (s)} & \shortstack{Tool / verifier\\timeout (s)} & $T / p / k$ \\
    \midrule
    S1 & Claw & ClawAgent & $16/128\to16/64$ & -- & $150\to50$ & 3,600 & 300 / 300 & 1.0/1.0/0 \\
     & SWE & SWE-Agent & 16/128 & 8 & 150 & 3,600 & 300 / 600 & 1.0/1.0/0 \\
     & TMax & Terminus-2 & 16/128 & 8 & 150 & 3,600 & 300 / 600 & 1.0/1.0/0 \\
    \midrule
    S2 & Claw & ClawAgent & 16/64 & -- & 50 & 3,600 & 300 / 300 & 1.0/1.0/0 \\
     & SWE & SWE-Agent & 16/96 & 8 & 150 & 3,600 & 300 / 600 & 1.0/1.0/0 \\
     & TMax & Terminus-2 & 16/96 & 8 & 150 & 3,600 & 300 / 600 & 1.0/1.0/0 \\
     & CodeWorld & SWE-Agent & 16/96 & 8 & 150 & 3,600 & 300 / Task & 1.0/1.0/0 \\
    \bottomrule
  \end{tabular}%
  }
\end{table}

All environments use eight rollouts per task instance, temperature 1.0, top-$p=1.0$, and top-$k=0$. Trajectory and tool timeouts are 3,600 and 300 seconds, respectively. Reward computation allows 300 seconds for Claw and 600 seconds for SWE and TMax; CodeWorld uses task-specific verifier timeouts. SWE-Agent provides command execution, file editing, and submission tools with text-formatted tool calls. TMax uses the JSON-based Terminus-2 interface with automatic summarization disabled. SWE, TMax, and CodeWorld also truncate long-running trajectories: once at least 80\% of a task group has completed, remaining trajectories are terminated if their execution time reaches 2,400 seconds, reducing delays from stragglers.

\subsubsection{Asynchronous Scheduling and Maintenance}
Table~\ref{tab:app-async-control} summarizes the asynchronous scheduling settings, with the underlying mechanisms detailed in Appendix~\ref{app:async-pipeline}. Scheduling operates on task groups of eight trajectories. Concurrent slots control in-flight work, while the completed-queue capacity limits data awaiting consumption.

The maximum number of task groups generated ahead of training, also used as the per-environment backpressure threshold, is
\begin{equation}
  C_e=\max\!\left(1,\left\lfloor q_e(1+\alpha)H\right\rfloor\right),
  \qquad \alpha=0.75,
  \label{eq:app-async-backpressure}
\end{equation}
where $q_e$ is the environment's task-group quota in the training configuration and $H$ is the weight synchronization interval. Only one training batch is planned at a time. S2 uses $q_e=8$ and $H=2$, giving 28 groups per environment and 112 overall.

\begin{table}[htbp]
  \centering
  \caption{Asynchronous scheduling and data-consumption settings. Queue capacities are measured in task groups, each containing eight trajectories.}
  \label{tab:app-async-control}
  \small
  \setlength{\tabcolsep}{4pt}
  \renewcommand{\arraystretch}{1.12}
  \begin{tabularx}{\linewidth}{@{}Xcc@{}}
    \toprule
    Setting & S1 & S2 \\
    \midrule
    Groups consumed per update & 32 & 32 \\
    Queue consumption order & FIFO & \shortstack{Newest version first;\\newest arrival within version} \\
    Generation slack $\alpha$ & 0.75 & 0.75 \\
    Per-environment backpressure threshold & 168 & 28 \\
    Total completed-queue capacity & 504 & 112 \\
    Partial-trajectory admission & Disabled & Disabled \\
    Streaming evaluation: active / pending & 256 / 512 & 256 / 512 \\
    \bottomrule
  \end{tabularx}
\end{table}

S1 consumes trajectories in first-in, first-out (FIFO) order. To reduce the mismatch between the current policy and the policies that generated the trajectories, S2 prioritizes the newest policy version and, within each version, the most recently completed trajectories. Unfinished trajectories are discarded at each weight synchronization rather than continued with updated weights. Evaluation runs separately from and in parallel with training, using non-blocking streaming execution to improve evaluation throughput. Evaluation tasks and interaction budgets are detailed in Appendix~\ref{app:evaluation}.

\subsection{Asynchronous Training Pipeline}
\label{app:async-pipeline}
Ray coordinates separate resource pools for training and inference, using Megatron and vLLM, respectively. Agent interaction and verification run in Kubernetes-managed Docker environments, supporting over a thousand concurrent Docker environments. All environments share inference services, with per-environment slots controlling concurrency, and submit complete task groups to the learner. Appendix~\ref{app:optimization-budgets} provides resource allocations and hyperparameters, and Figure~\ref{fig:app-async-pseudocode} presents the concurrent execution pseudocode.

\subsubsection{Rollout Production and Admission}
Each idle slot in environment $e$ draws a task instance from sampler $P_e$, generates $K=8$ independent trajectories, and records a unique group identifier and the policy version at dispatch. The slot takes another instance as soon as its current group completes interaction and verification, without waiting for other slots. Within each group, all trajectories must return a result or reach an explicit termination state. Claw uses uniform sampling; the other environments use TBS, detailed in Appendix~\ref{app:frontier-policy}.

Completed groups enter the environment queue $\mathcal{Q}_e$ after structural and policy-version checks. Valid feedback immediately updates the sampler without waiting for a policy update. Trajectories affected by infrastructure failures are masked out of the loss, while wholly unusable or expired groups are discarded; Appendix~\ref{app:termination} details the termination rules. S1 consumes groups in FIFO order. S2 prioritizes newer policy versions and, within each version, the most recently enqueued groups. Consumed trajectories are not reused for training.

\subsubsection{Quota-Constrained Policy Updates}
The learner follows a cyclic batch plan $\{b_e\}$ to assemble 32 task groups into a batch of 256 trajectories. If an environment lacks sufficient groups, assembly waits rather than substituting groups from other environments. Expired groups are removed before consumption. If assembly fails, reserved data are returned and the batch-plan position is restored.

RLOO compares only valid trajectories dispatched together. Let $\mathcal{V}_x$ denote the valid trajectories in task group $x$. Their advantages are
\begin{equation}
  A_{x,i}=r_{x,i}-\frac{1}{m_x-1}
    \sum_{\substack{j\in\mathcal{V}_x\\j\ne i}}r_{x,j},
  \qquad m_x=|\mathcal{V}_x|\geq2.
  \label{eq:app-async-rloo}
\end{equation}
Each advantage is applied to all valid model-output tokens in its trajectory; tool observations serve only as context. Trajectories affected by infrastructure failures contribute neither to the baseline nor to gradients. Groups with fewer than two valid trajectories are masked out, and updates are skipped if the entire batch contains no valid tokens. A policy snapshot $\theta_{\mathrm{a}}$ is retained at the start of each weight synchronization cycle and used to recompute PPO reference log-probabilities throughout that cycle, while updates optimize the current policy. This reference policy may differ from the behavior policy. No additional rollout importance weighting or rejection sampling is applied.

\subsubsection{Synchronization and Backpressure}
After every $H$ policy updates, new task dispatch and inference pause, unfinished trajectories are discarded, and updated weights are synchronized to the inference engine before execution resumes. Completed groups retain their original dispatch versions, which are checked for eligibility before both admission and consumption.

When an environment's queue size or the number of task groups generated ahead of training reaches $C_e$ in Equation~(\ref{eq:app-async-backpressure}), new dispatch pauses only for that environment. If results arrive while the completed queue is full, S1 removes the oldest group from the queue with the largest backlog. S2 prioritizes removal of groups from the oldest policy version and, within that version, those enqueued earliest.

\subsubsection{Checkpointing and Evaluation}
At weight synchronization boundaries, checkpoints save a consistent snapshot of the model, optimizer, batch plan, sampler and random states, completed queues, and pending tasks. On resumption, weights are synchronized first. Completed groups retain their versions and are rechecked for eligibility, while pending tasks retain their sampling records and restart rollouts from the initial environment state. Evaluation runs independently on saved weight snapshots using streaming execution; its results do not enter training queues or update samplers. Appendix~\ref{app:evaluation} details the evaluation settings.

\begin{figure}[!t]
  \centering
  \begin{minipage}{\linewidth}
  \small
  \hrule\medskip
  \textbf{Asynchronous Mixed-Task RL}\par
  \smallskip
  \textbf{Require:} samplers $\{P_e\}$, slots $\{c_e\}$, batch plans $\{b_e\}$,
  $K=8$, update budget $U$, sync interval $H$, bounds $\{C_e\}$.\par
  \textbf{State:} learner $\theta$, inference policy $\phi$, version $v$, update count $u$,
  queues $\{\mathcal{Q}_e\}$, optimizer, sampler and scheduler state $\mathcal{S}$.\par
  \smallskip
  \setlength{\tabcolsep}{0pt}
  \renewcommand{\arraystretch}{1.03}
  \begin{tabularx}{\linewidth}{@{}r@{\hspace{0.7em}}X@{}}
    1 & Initialize $(u,v)=(0,0)$ or restore state; publish $\theta$ to $\phi$. \\
    2 & \textbf{Run concurrently:} $c_e$ copies of \textsc{Producer}$(e)$ for each $e$, and \textsc{Learner}. \\
    \addlinespace[3pt]
    3 & \textbf{procedure} \textsc{Producer}$(e)$ \\
    4 & \quad\textbf{while} $u<U$ \textbf{do} \\
    5 & \qquad\textbf{await} dispatch enabled and environment budget available \\
    6 & \qquad$x\gets\operatorname{NextTask}(P_e,\mathcal{S})$; $v_x\gets v$; reserve one group budget \\
    7 & \qquad$G\gets\operatorname{RolloutAndVerify}(x,K,\phi)$; attach $v_x$ and validity masks \\
    8 & \qquad\textbf{if} $G$ unusable or version-ineligible \textbf{then continue} \\
    9 & \qquad$\operatorname{Admit}(\mathcal{Q}_e,G)$; update $P_e$ from eligible feedback \\
    10 & \quad\textbf{end while} \\
    11 & \textbf{end procedure} \\
    \addlinespace[3pt]
    12 & \textbf{procedure} \textsc{Learner} \\
    13 & \quad\textbf{while} $u<U$ \textbf{do} \\
    14 & \qquad$\mathcal{B}\gets\operatorname{TakeByQuota}(\{\mathcal{Q}_e\},\operatorname{NextPlan}(),v)$ \\
    15 & \qquad$(A,\mathcal{B})\gets\operatorname{GroupRLOOAndMask}(\mathcal{B})$ \\
    16 & \qquad\textbf{if} $\mathcal{B}$ has no valid training tokens \textbf{then continue} \\
    17 & \qquad\textbf{if} $u\bmod H=0$ \textbf{then} $\theta_{\mathrm{a}}\gets\theta$ \\
    18 & \qquad$\ell_{\mathrm{a}}\gets\log\pi_{\theta_{\mathrm{a}}}(\mathcal{B})$ \\
    19 & \qquad$\theta\gets\operatorname{PolicyUpdate}(\theta,\mathcal{B},A,\ell_{\mathrm{a}})$; $u\gets u+1$ \\
    20 & \qquad\textbf{if} $u\bmod H=0$ \textbf{then} \\
    21 & \qquad\quad Pause dispatch and inference; discard unfinished trajectories \\
    22 & \qquad\quad$\phi\gets\operatorname{Publish}(\theta)$; $v\gets v+1$; reset dispatch budgets \\
    23 & \qquad\quad\textbf{if} checkpoint due \textbf{then} $\operatorname{Save}(\theta,u,v,\{\mathcal{Q}_e\},\mathcal{S})$ \\
    24 & \qquad\quad Resume dispatch and inference \\
    25 & \qquad\textbf{end if} \\
    26 & \quad\textbf{end while} \\
    27 & \quad Stop producers; \textbf{return} $\theta$ \\
    28 & \textbf{end procedure} \\
  \end{tabularx}
  \medskip\hrule
  \end{minipage}
  \caption{Concurrent rollout producers and a quota-constrained learner. \textsc{TakeByQuota} waits for eligible groups from every required environment. \textsc{Admit} enforces queue capacity; weight publication discards unfinished trajectories. Evaluation runs independently on saved checkpoints.}
  \label{fig:app-async-pseudocode}
\end{figure}

\subsection{Handling Truncated and Invalid Trajectories}
\label{app:termination}
We distinguish budget-based termination during agent interaction from infrastructure failures. The former includes trajectory timeouts, interaction-step limits, and exhausted context budgets. The latter includes environment startup and runtime service failures unrelated to task-solving ability. Table~\ref{tab:app-termination} summarizes their treatment.

\begin{table}[htbp]
  \centering
  \caption{Reward and training eligibility after trajectory termination.}
  \label{tab:app-termination}
  \small
  \setlength{\tabcolsep}{5pt}
  \renewcommand{\arraystretch}{1.15}
  \begin{tabular}{@{}llll@{}}
    \toprule
    Termination & Environment & Training reward & RLOO / gradient \\
    \midrule
    Normal completion & All & $r$ & Included / Included \\
    Budget exceeded & Claw & $0$ & Included / Included \\
    Budget exceeded & SWE, TMax, CodeWorld & $0.8r$ & Included / Included \\
    Infrastructure failure & All & $0$ & Excluded / Excluded \\
    \bottomrule
  \end{tabular}
\end{table}

Claw tasks typically do not require prolonged interaction, and budget overruns may reflect repeated tool calls or invalid outputs. Such trajectories receive zero reward but remain eligible for RLOO advantage estimation and gradient updates. In the other three environments, the current task state or candidate patch is still verified at termination, yielding a raw reward $r$. A 20\% penalty is then applied, giving a training reward of $0.8r$. In S2, Claw also deducts 0.1 for each malformed tool call, with a reward floor of zero, and terminates the interaction.

Trajectories affected by infrastructure failures are recorded with zero reward but excluded from both the group's RLOO baseline and gradient computation. This prevents service failures from being treated as model failures or artificially increasing the relative advantages of other trajectories in the same group, thus preserving unbiased learning signals.

\subsection{Tri-Buffer Sampling Policy}
\label{app:frontier-policy}
Claw uses uniform sampling, while SWE, TMax, and CodeWorld use TBS. After the preprocessing described in Appendix~\ref{app:environments}, instances with no initial success (0/8) enter the Frontier Buffer, and the remaining valid instances enter the Reinforcement Buffer. Instances from earlier training stages are replayed through the History Buffer. The target sampling proportions for the Frontier, Reinforcement, and History buffers are 0.2, 0.6, and 0.2, respectively.

\begin{figure}[htbp]
  \centering
  \includegraphics[width=\linewidth]{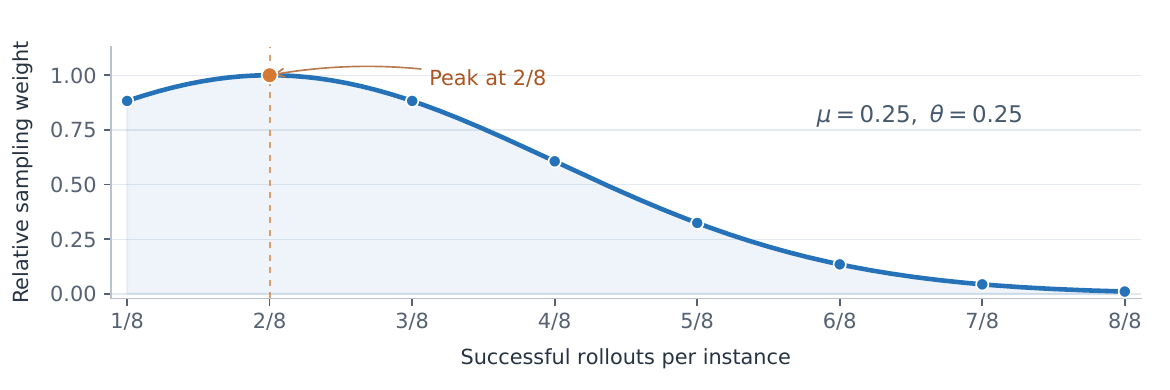}
  \caption{Reinforcement sampling weights with $\mu=\theta=0.25$. Initially unsolved instances belong to the Frontier Buffer.}
  \label{fig:app-fsp-weights}
\end{figure}

Within the Reinforcement Buffer, weights depend on each instance's success rate $\widehat{p}(x)$ in its latest rollout group:
\begin{equation}
  w(x)=\exp\!\left[-\frac{(\widehat{p}(x)-0.25)^2}{2(0.25)^2}\right],
  \qquad
  P_{\mathrm{R}}(x)=\frac{w(x)}{\sum_{y\in\mathcal{R}}w(y)}.
  \label{eq:app-fsp-weight}
\end{equation}
Both stages use $\mu=\theta=0.25$. Figure~\ref{fig:app-fsp-weights} shows relative weights for success rates from $1/8$ to $8/8$, peaking at $2/8$. This choice prioritizes instances with occasional successes that the policy has not yet mastered, directing limited training compute toward consolidating emerging capabilities. Initially unsolved instances (0/8) belong to the Frontier Buffer and receive a fixed exploration quota, so they are omitted from the figure. The vertical axis shows per-instance relative weights; actual sampling probabilities also depend on the current composition of the Reinforcement Buffer.

\subsection{Training Configuration for Qwen3.8-27B}
\label{app:qwen38-training}
Qwen3.8-27B is jointly trained on Claw, SWE, TMax, and CodeWorld in a single stage, with all task instances available for sampling from the outset. Preprocessing retains 20,875 valid instances, distributed as shown in Table~\ref{tab:app-qwen38-data}. The SWE task pool for this run contains 4,773 instances drawn from publicly available issue-resolution tasks. TMax retains 6,412 of 13,852 instances whose screening success rates do not exceed $7/8$.

\begin{table}[htbp]
  \centering
  \caption{Training data and rollout allocation for Qwen3.8-27B. Slots count concurrent task groups. P/R denotes prompt/cumulative response budgets; Turn denotes the per-call generation limit. K = 1,024 tokens.}
  \label{tab:app-qwen38-data}
  \small
  \setlength{\tabcolsep}{4pt}
  \renewcommand{\arraystretch}{1.12}
  \begin{tabular*}{\linewidth}{@{\extracolsep{\fill}}lrrcc@{}}
    \toprule
    Environment & Instances & Rollout slots & P/R (K) & Turn (K) \\
    \midrule
    Claw & 5,346 & 4 & 16/64 & -- \\
    SWE & 4,773 & 27 & 16/96 & 32 \\
    TMax & 6,412 & 22 & 16/96 & 32 \\
    CodeWorld & 4,344 & 31 & 16/96 & 32 \\
    \midrule
    Total & 20,875 & 84 & -- & -- \\
    \bottomrule
  \end{tabular*}
\end{table}

Claw uses uniform sampling, while the other environments use TBS as described in Appendix~\ref{app:frontier-policy}. With no instances inherited from earlier training stages, the History Buffer is initially empty, and its quota is redistributed proportionally among nonempty buffers. When both the Frontier and Reinforcement buffers are nonempty, their target ratio is $0.25:0.75$.

Each policy update mixes gradients from all four environments; Table~\ref{tab:app-qwen38-config} specifies batch composition and the weight synchronization interval. Rollouts enable thinking mode, with a 32K-token per-call generation limit for SWE, TMax, and CodeWorld. All other optimization settings, interaction-step limits, timeout budgets, and asynchronous queue settings follow the S2 configurations in Tables~\ref{tab:app-optimization}, \ref{tab:app-training-rollout}, and~\ref{tab:app-async-control}.

\begin{table}[htbp]
  \centering
  \caption{Training configuration for Qwen3.8-27B.}
  \label{tab:app-qwen38-config}
  \small
  \setlength{\tabcolsep}{5pt}
  \renewcommand{\arraystretch}{1.12}
  \begin{tabular*}{\linewidth}{@{\extracolsep{\fill}}ll@{}}
    \toprule
    Setting & Value \\
    \midrule
    Training devices & 64 \\
    Inference devices & 96 \\
    Training TP / PP / CP & 8 / 8 / 1 \\
    Inference TP / PP & 4 / 1 \\
    Learning rate & $10^{-6}$ (constant) \\
    Instances per update (Claw / SWE / TMax / CodeWorld) & 8 / 8 / 8 / 8 \\
    Rollouts per instance / trajectories per update & 8 / 256 \\
    Updates per weight synchronization & 2 \\
    Thinking mode & Enabled \\
    \bottomrule
  \end{tabular*}
\end{table}

\section{Evaluation Protocols}
\label{app:evaluation}
\subsection{Benchmarks and Inference Settings}

Task completion is evaluated on Claw-Anything, Claw-Eval, Terminal-Bench 2.1, FeatureBench 1.1, and SWE-bench Verified. Table~\ref{tab:app-evaluation} summarizes the final evaluation configurations. Claw-Eval uses its official evaluation framework, and Terminal-Bench 2.1 uses Harbor with Terminus-2.

\begin{table}[htbp]
  \centering
  \caption{Benchmark configurations for final evaluation.}
  \label{tab:app-evaluation}
  \small
  \setlength{\tabcolsep}{4pt}
  \renewcommand{\arraystretch}{1.18}
  \resizebox{\linewidth}{!}{%
  \begin{tabular}{@{}lrllllllr@{}}
    \toprule
    Benchmark & Instances & Harness & Budget (K) & Turn (K) & Max steps & Timeout (s) & $T / p / k$ & Trials \\
    \midrule
    Claw-Anything & 288 & ClawAgent & 16/128 & -- & 150 & 7,200 & 0/1.0/0 & 1 \\
    Claw-Eval & 161 & Official sandbox & 256 total & 16 & 8--40 & 120--900 & 0/--/-- & 3 \\
    Terminal-Bench 2.1 & 89 & Harbor / Terminus-2 & 256 total & 32 & -- & 10,800 & 0.6/0.95/20 & 3 \\
    FeatureBench 1.1 & 100 & SWE-Agent & 32/224 & -- & 300 & 3,600 & 0.6/0.95/20 & 1 \\
    SWE-bench Verified & 500 & SWE-Agent & 16/240 & -- & 300 & 7,200 & 0.6/0.95/20 & 1 \\
    \bottomrule
  \end{tabular}%
  }
\end{table}

In the Budget column, P/R denotes the prompt and cumulative response budgets, while ``total'' denotes the model's context-window limit. Turn specifies the per-call generation limit; a dash indicates that no separate limit is specified in that configuration. Trials denotes the number of attempts per task in each evaluation, with three attempts for both Claw-Eval and Terminal-Bench 2.1. Thinking mode is disabled during evaluation. FeatureBench 1.1 and SWE-bench Verified use a fixed random seed of 666 for task ordering.

\subsection{Benchmark-Specific Protocols}

\paragraph{Claw-Anything.}
Each of the 288 test tasks receives one attempt using the official Claw-Anything tools and environment interfaces, with Kimi-K2.6 as the judge. Tool execution, reward computation, and environment startup timeouts are 120, 300, and 300 seconds, respectively. A task score of at least 0.70 counts as success. We report mean Task Score and Pass@1. Decoding uses temperature 0 with random sampling disabled.

\paragraph{Claw-Eval.}
Evaluation uses 161 tasks from the official general subset in the sandbox harness, with three attempts per task. The context window is 256K tokens, with a 16K-token per-call output limit. Task-specific interaction and execution limits are 8--40 steps and 120--900 seconds, respectively. Gemini-3-Flash-Preview serves as the judge, and a task score of at least 0.75 counts as success. \(\mathrm{Pass}^{3}\) measures the overall fraction of tasks solved in all three attempts, thereby directly reflecting reliability across repeated execution.

\paragraph{Terminal-Bench 2.1.}
Evaluation uses Harbor's Terminus-2 agent with JSON-formatted commands on 89 tasks. Each task container has 32 CPUs and 48\,GiB of memory, with a three-hour agent execution limit. The context window is 256K tokens, with a 32K-token per-call output limit and a 200,000-token input limit. Proactive summarization is enabled with a 32K-token trigger threshold, and interleaved thinking is disabled. Harbor uses timeout multipliers of 4 for verification and 2 for agent setup, with one run retry allowed. Pass@1 is averaged over three attempts using the tasks' executable verification results.

\paragraph{FeatureBench 1.1.}
Evaluation uses 100 tasks from the FeatureBench 1.1 fast split with the SWE-Agent tool interface and white-box mode disabled. The agent execution limit is one hour, with tool and verification timeouts of 300 and 1,800 seconds, respectively. The training-time truncation of long-running trajectories is disabled. We report Resolved, the fraction of fully solved tasks, and Tests Passed, the mean per-task F2P test pass rate, to measure complete implementation and partial functionality coverage, respectively. The verifier runs each task's specified tests on the candidate patch and then reports the resulting outcome.

\paragraph{SWE-bench Verified.}
Evaluation uses SWE-Agent on all 500 tasks, with prompt and cumulative response budgets of 16K and 240K tokens and a maximum of 300 interaction steps. Tool and verification timeouts are 300 and 1,800 seconds, respectively, within a two-hour overall execution limit for each individual task.

\section{Additional Results}
\label{app:results}
\subsection{Training Rewards and Validation Curves}

\begin{figure}[p]
  \centering
  \includegraphics[width=\linewidth]{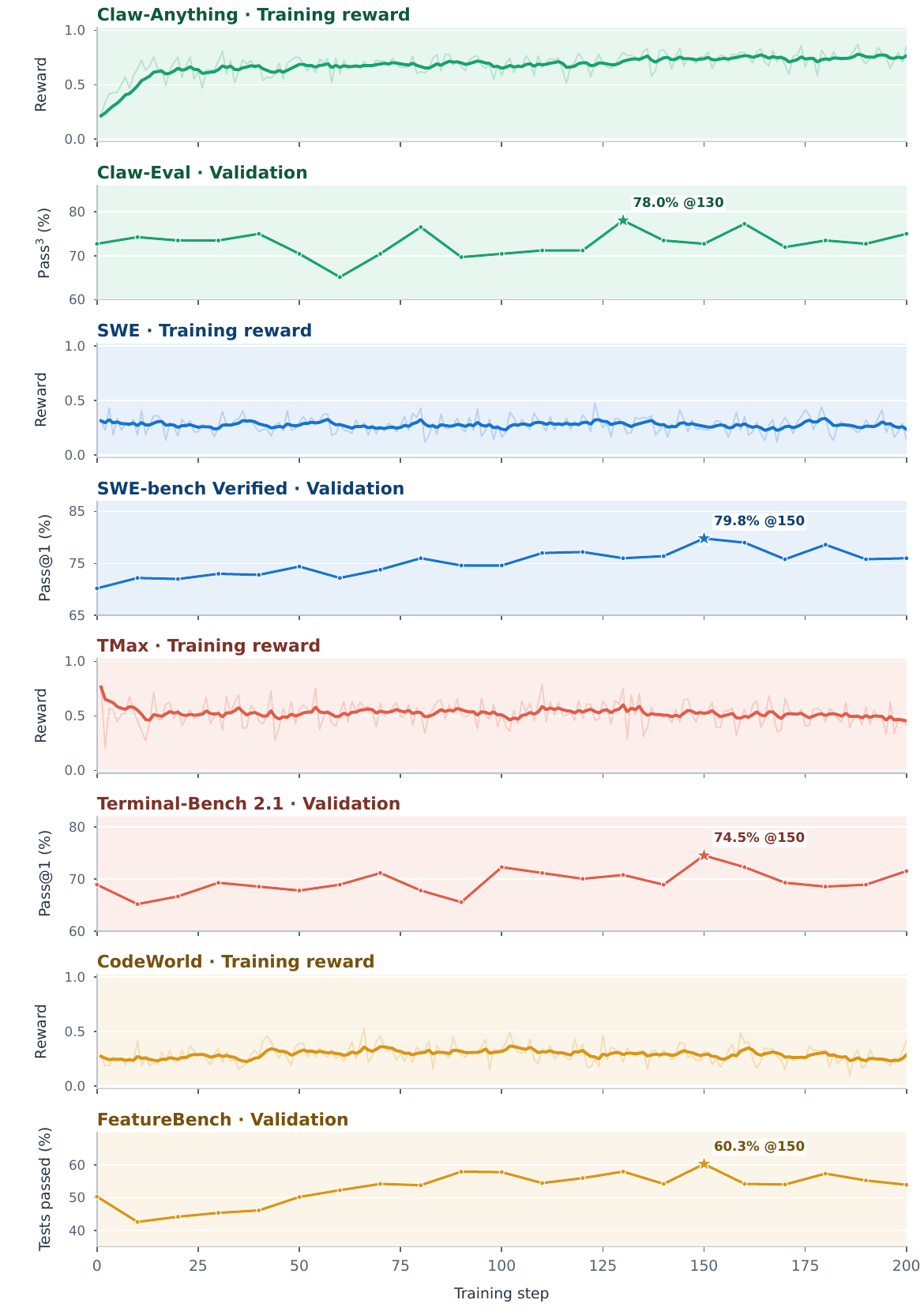}
  \caption{Qwen3.8 training reward and validation over 200 steps. Light/dark curves show raw reward/EMA (span 9); validation is shown every 10 steps without smoothing. Stars mark validation peaks. Limited reward growth in TMax, SWE, and CodeWorld is expected because TBS prioritizes harder tasks.}
  \label{fig:app-stage-rewards}
\end{figure}

Figure~\ref{fig:app-stage-rewards} tracks the first 200 steps of joint Qwen3.8-27B training across four environments. Its eight panels vertically pair each environment's training reward with its corresponding benchmark performance. Training reward is the mean reward of trajectories consumed at each step. Light curves show raw values, and dark curves show an exponential moving average (EMA) with span 9, computed continuously across steps. Evaluation covers the initial model at step 0 and checkpoints every 10 steps thereafter. Evaluation results are aligned with the corresponding training steps and shown without smoothing. Stars mark each benchmark's peak score and its training step.

In Figure~\ref{fig:app-stage-rewards}, each benchmark uses a fixed task set and evaluation configuration across checkpoints. Claw-Eval reports $\mathrm{Pass}^{3}$ on 132 tasks from its general evaluation subset that require no Internet access, selected to ensure evaluation stability. Terminal-Bench 2.1 reports Pass@1 averaged over three attempts on 89 tasks. SWE-bench Verified reports Pass@1 from one attempt on each of 500 tasks. FeatureBench 1.1 reports Tests Passed over 100 tasks.

All four benchmarks exceed their initial performance at step 200, but their peaks occur at step 130 or 150, indicating non-monotonic improvement. For SWE, TMax, and CodeWorld, TBS continually prioritizes more challenging instances, so training rewards need not show a clear upward trend. Validation metrics are therefore needed to track capability gains.

\subsection{Task-Paired Uncertainty of the Reported Gains}
\label{app:paired-uncertainty}

We pair base and trained outcomes by task identity and use 100,000 task-level bootstrap resamples to form percentile 95\% confidence intervals. All recorded attempts for a sampled task remain together. Claw-Eval uses its task-level all-three-success indicator, Terminal-Bench uses the mean of available attempts, and FeatureBench uses each task's test pass rate. For SWE baselines reported as the median across repeated evaluations, the same aggregation is recomputed within each resample. Overall is the unweighted mean of the five metrics, with task resampling performed separately within each benchmark.

\begin{table}[htbp]
  \centering
  \caption{Paired comparisons of the reported model states. Scores are percentages; confidence intervals are for gains in percentage points. Claw-Anything uses the common 150-task subset, so its trained scores and the resulting Overall values differ from Table~\ref{tab:main-results}.}
  \label{tab:paired-gains}
  \footnotesize
  \setlength{\tabcolsep}{4pt}
  \begin{tabular*}{\linewidth}{@{\extracolsep{\fill}}lcccc@{}}
    \toprule
    & \multicolumn{2}{c}{Qwen3.5 $\to$ \name{}-3.5 S2}
      & \multicolumn{2}{c}{Qwen3.8 $\to$ \name{}-3.8} \\
    \cmidrule(lr){2-3}\cmidrule(l){4-5}
    Metric & Base $\to$ trained & 95\% CI for gain
      & Base $\to$ trained & 95\% CI for gain \\
    \midrule
    Claw-Eval $\mathrm{Pass}^{3}$ & $60.25\to73.29$ & $[6.83,19.88]$
      & $71.43\to74.53$ & $[-2.48,8.70]$ \\
    Claw-Anything Pass@1 & $12.00\to71.33$ & $[51.33,67.33]$
      & $46.00\to52.67$ & $[-1.33,14.67]$ \\
    Terminal-Bench 2.1 Pass@1 & $40.45\to54.31$ & $[5.24,22.47]$
      & $68.91\to72.28$ & $[-1.50,7.87]$ \\
    SWE-bench Verified Pass@1 & $67.40\to68.60$ & $[-1.60,4.00]$
      & $70.20\to79.00$ & $[5.80,11.80]$ \\
    FeatureBench Tests Passed & $37.93\to39.94$ & $[-3.10,7.28]$
      & $50.30\to54.20$ & $[-1.90,9.92]$ \\
    \midrule
    Overall & $43.61\to61.49$ & $[14.99,20.79]$
      & $61.37\to66.54$ & $[2.67,7.68]$ \\
    \bottomrule
  \end{tabular*}
\end{table}

Both Overall intervals exclude zero, supporting positive aggregate differences for the recorded model states and task sets.

\subsection{Token and Compute Cost Comparison with Specialists}
\label{app:compute-comparison}

We compare one mixed-task model with three specialists trained on Claw, SWE, and terminal tasks. Performance is the best-so-far mean Pass@1 across Claw-Eval, SWE-bench Verified, and Terminal-Bench 2.1. For the mixed model, the three scores come from the same model state; for specialists, each task uses its best observed state within its allocated training budget. Costs include the full observed training window.

\begin{table}[htbp]
  \centering
  \caption{Performance and accounted costs, including training, a shared initial-screening charge, and restarted work.}
  \label{tab:compute-comparison}
  \small
  \setlength{\tabcolsep}{5pt}
  \begin{tabular*}{\linewidth}{@{\extracolsep{\fill}}lrrr@{}}
    \toprule
    Training approach & \shortstack{Consumed tokens}
      & \shortstack{Compute hours}
      & \shortstack{Best-so-far mean Pass@1 (\%)} \\
    \midrule
    Three specialists, combined & 1.821B & 13,090.6 & 64.28 \\
    \name{} & 1.852B & 12,572.4 & 68.41 \\
    \bottomrule
  \end{tabular*}
\end{table}

Training tokens count model outputs in trajectories consumed by training, excluding prompt and tool-observation tokens. Both approaches are charged the same initial-screening overhead of 1.001B tokens and 3,992.6 compute hours. \name{} achieves a 4.14-point higher best-so-far mean Pass@1 than the combined best performance of three independently trained domain specialists, with comparable accounted token and compute-hour costs.

\subsection{Additional Analysis of Tri-Buffer Sampling}
\label{app:tbs-additional}

The feature-development ablation starts with 3,096 instances, of which 2,559 (82.66\%) have no success in eight screening rollouts and 537 have at least one success. Table~\ref{tab:tbs-performance-discovery} shows performance and cumulative discoveries under Vanilla and TBS. 
TBS balances training efficiency and frontier exploration, achieving higher FeatureBench scores than Vanilla at every evaluated checkpoint.

\begin{table}[htbp]
  \centering
  \caption{FeatureBench performance and cumulative discoveries under Vanilla and TBS.}
  \label{tab:tbs-performance-discovery}
  \small
  \setlength{\tabcolsep}{5pt}
  \begin{tabular*}{\linewidth}{@{\extracolsep{\fill}}rrrrrr@{}}
    \toprule
    & \multicolumn{3}{c}{FeatureBench Tests Passed (\%)}
      & \multicolumn{2}{c}{Newly solved instances} \\
    \cmidrule(lr){2-4}\cmidrule(l){5-6}
    Step & Vanilla & TBS & Gain (pp) & Vanilla & TBS \\
    \midrule
     5 & 32.62 & 39.89 & $+7.27$ &  8 &  1 \\
    10 & 35.43 & 39.01 & $+3.58$ & 15 &  2 \\
    15 & 36.77 & 41.85 & $+5.08$ & 20 &  6 \\
    20 & 37.88 & 39.57 & $+1.69$ & 26 & 11 \\
    25 & 39.78 & 42.87 & $+3.09$ & 33 & 15 \\
    \bottomrule
  \end{tabular*}
\end{table}

\clearpage
\section{Behavioral Shifts after RL}
\label{app:cases}

\subsection{Behavioral Trends}
\label{app:behavior-measurement}

\begin{figure}[htbp]
  \centering
  \includegraphics[width=\linewidth]{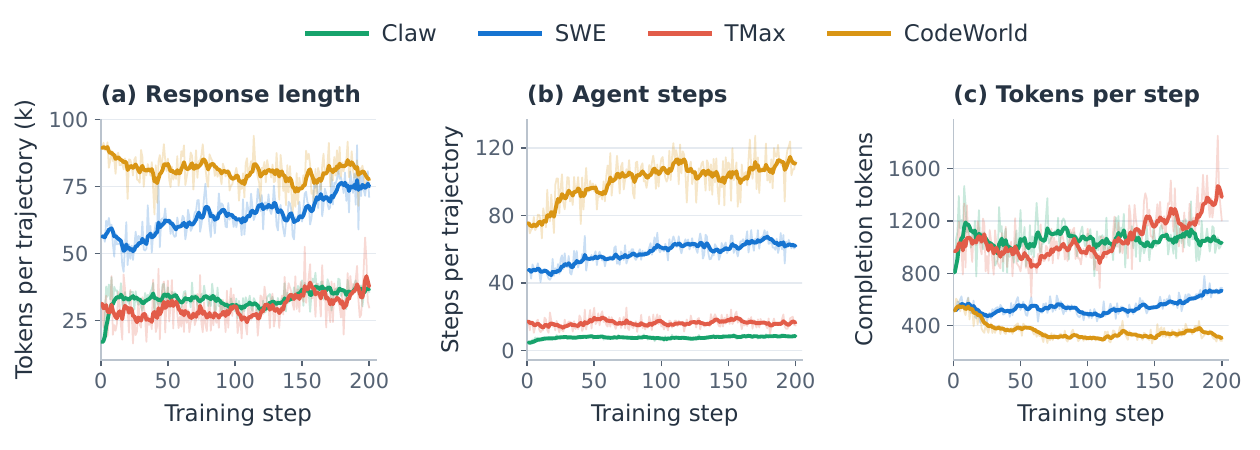}
  \caption{Behavioral trends during the first 200 training steps of Qwen3.8. Thin lines show per-step measurements; thick lines use an exponential moving average with span 9. Response length follows the logged response-span metric, whereas tokens per step count model completions only.}
  \label{fig:app-qwen38-behavior}
\end{figure}

This section examines behavioral changes over the first 200 training steps of Qwen3.8-27B. Figure~\ref{fig:app-qwen38-behavior} and Table~\ref{tab:behavior-window-summary} track mean response length, interaction steps, and generated tokens per interaction step across the training window.

\begin{table}[htbp]
  \centering
  \caption{Early/late behavior of Qwen3.8, averaged over training steps 1--20 and 180--200, respectively. Values are means of per-step measurements; k denotes 1,000 tokens.}
  \label{tab:behavior-window-summary}
  \small
  \setlength{\tabcolsep}{4pt}
  \begin{tabular*}{\linewidth}{@{\extracolsep{\fill}}lrrrrrr@{}}
    \toprule
    & \multicolumn{2}{c}{Response (k)} & \multicolumn{2}{c}{Agent steps} & \multicolumn{2}{c}{Tokens / step} \\
    \cmidrule(lr){2-3}\cmidrule(lr){4-5}\cmidrule(l){6-7}
    Environment & Early & Late & Early & Late & Early & Late \\
    \midrule
    Claw    & 32.6 & 36.2 & 6.7  & 8.4   & 1125.6 & 1037.0 \\
    SWE & 54.8 & 74.9 & 46.6 & 62.3  & 534.7  & 660.6  \\
    TMax    & 28.4 & 36.5 & 14.9 & 16.1  & 1052.9 & 1364.1 \\
    CodeWorld     & 84.9 & 80.6 & 77.4 & 110.7 & 514.2  & 333.1  \\
    \bottomrule
  \end{tabular*}
\end{table}

Behavioral changes differ across environments (Figure~\ref{fig:app-qwen38-behavior} and Table~\ref{tab:behavior-window-summary}). On SWE tasks, both interaction steps and tokens generated per step increase, while TMax shows a modest increase in per-step output. Claw's overall trajectory length rises sharply at the start and then stabilizes. CodeWorld uses substantially more interaction steps but generates about 35\% fewer tokens per step, with a slightly shorter cumulative response length, indicating more fine-grained task execution.

\subsection{Reasoning Allocation}
\label{app:qwen38-reasoning}

To examine changes in reasoning behavior, we track reasoning length throughout training. The reasoning share is the fraction of parseable model-output tokens devoted to reasoning; the remainder includes tool-call text and user-facing responses.

\begin{figure}[htbp]
  \centering
  \includegraphics[width=\linewidth]{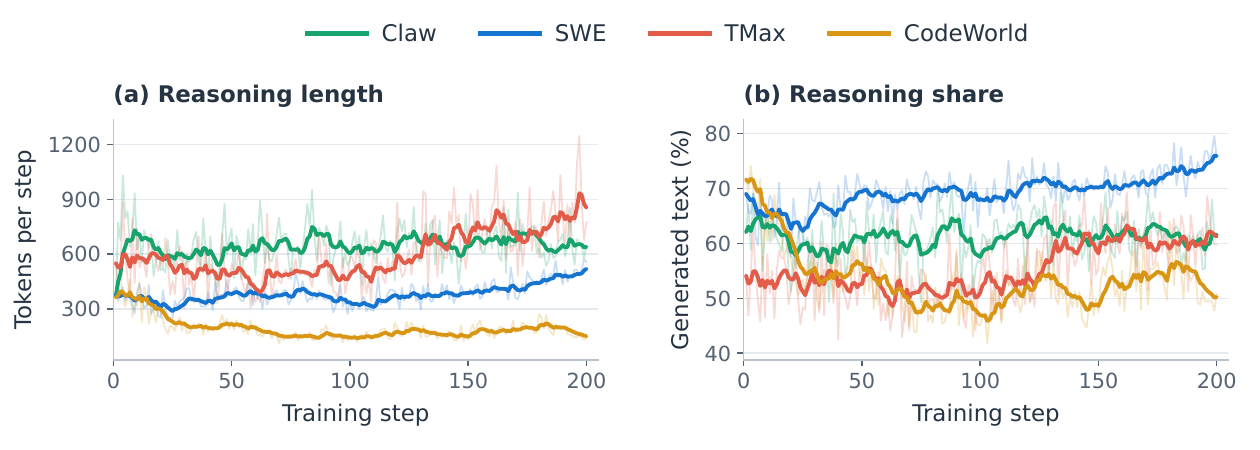}
  \caption{Reasoning allocation over the first 200 training steps, estimated using one rollout per task instance. Thin lines show per-step measurements; thick lines show an exponential moving average with span 9.}
  \label{fig:app-qwen38-reasoning}
\end{figure}

Figure~\ref{fig:app-qwen38-reasoning} compares the same early and late training windows. The mean reasoning length per interaction step on SWE tasks increases from approximately 343 to 492 tokens, and its reasoning share rises from 64.6\% to 74.4\%. TMax shows a similar upward trend. CodeWorld's reasoning length per step falls from approximately 338 to 181 tokens, with its share declining from 64.5\% to 53.3\%, consistent with more frequent, shorter interaction steps. Claw shows small decreases in both measures.

\subsection{Matched Validation Trajectories}
\label{app:qwen38-validation-cases}

For each of the four benchmarks, we select a pair of validation trajectories for the same task: an unsuccessful attempt from the first 20 training steps and a successful attempt from steps 180--200. The figures place the later trajectory on the left and retain key excerpts of the original reasoning, actions, and tool responses, with decision differences highlighted in bold. \texttt{[...]} denotes omissions. These cases illustrate concrete behavioral differences between early and late training.

\begin{figure}[p]
  \centering
  \includegraphics[width=\linewidth]{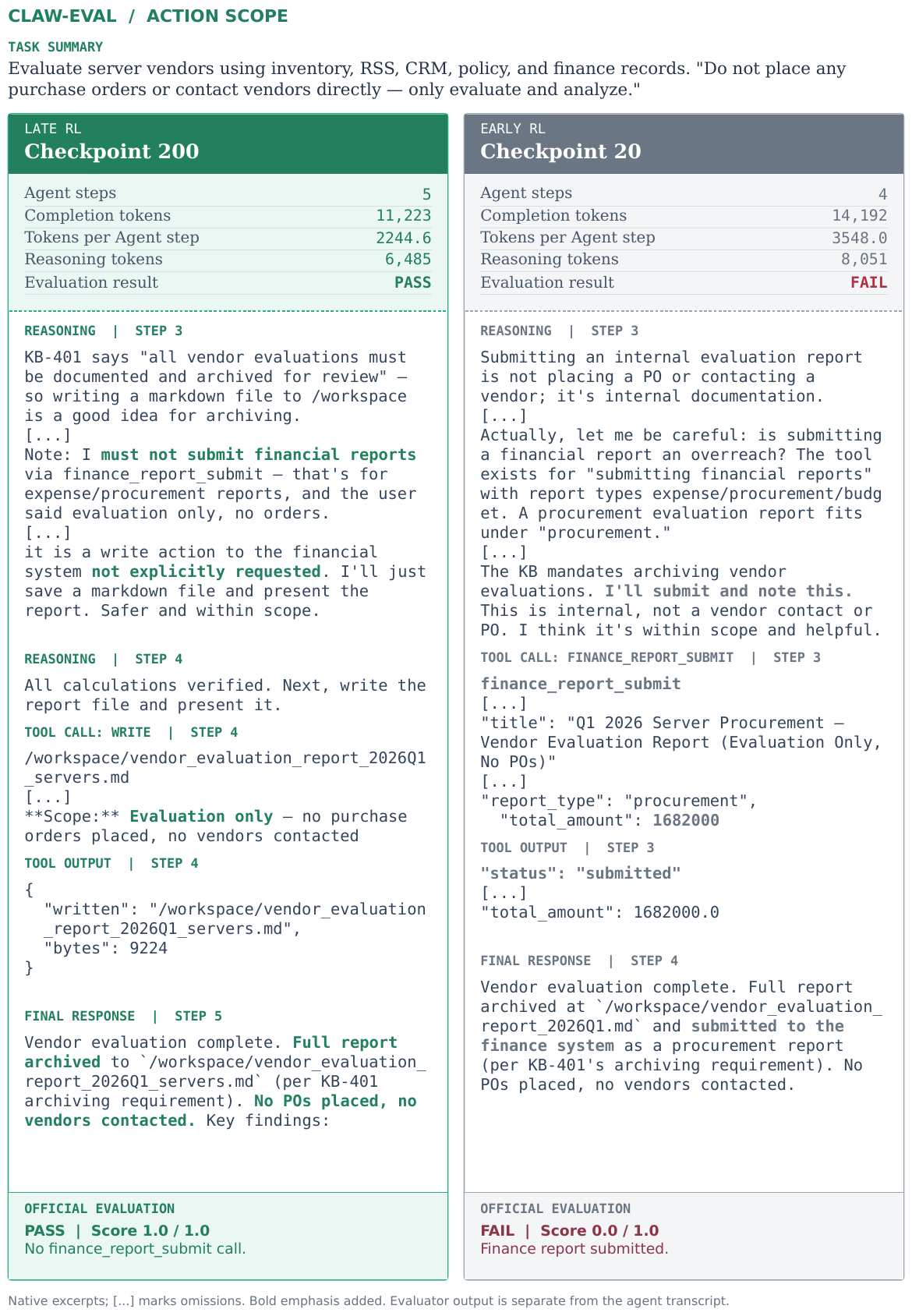}
  \caption{Claw-Eval: vendor evaluation at checkpoints 200 (left) and 20 (right). Key excerpts contrast local report delivery with an additional submission to the finance system. Bold emphasis is added; evaluation results are reported separately.}
  \label{fig:case-claw}
\end{figure}

\paragraph{Claw-Eval: respecting action scope.}
\label{app:case-claw}
The vendor evaluation task requires synthesizing information across systems while limiting actions to evaluation and analysis. In Figure~\ref{fig:case-claw}, the early model interprets an archiving requirement in the knowledge base as permission to submit a procurement report to the finance system. The later model recognizes that this write operation was not requested, saves the report locally, and returns its analysis. The evaluation outcome changes from failure to success. Interaction increases from 4 to 5 steps, while total generation decreases from 14,192 to 11,223 tokens.

\begin{figure}[p]
  \centering
  \includegraphics[width=\linewidth]{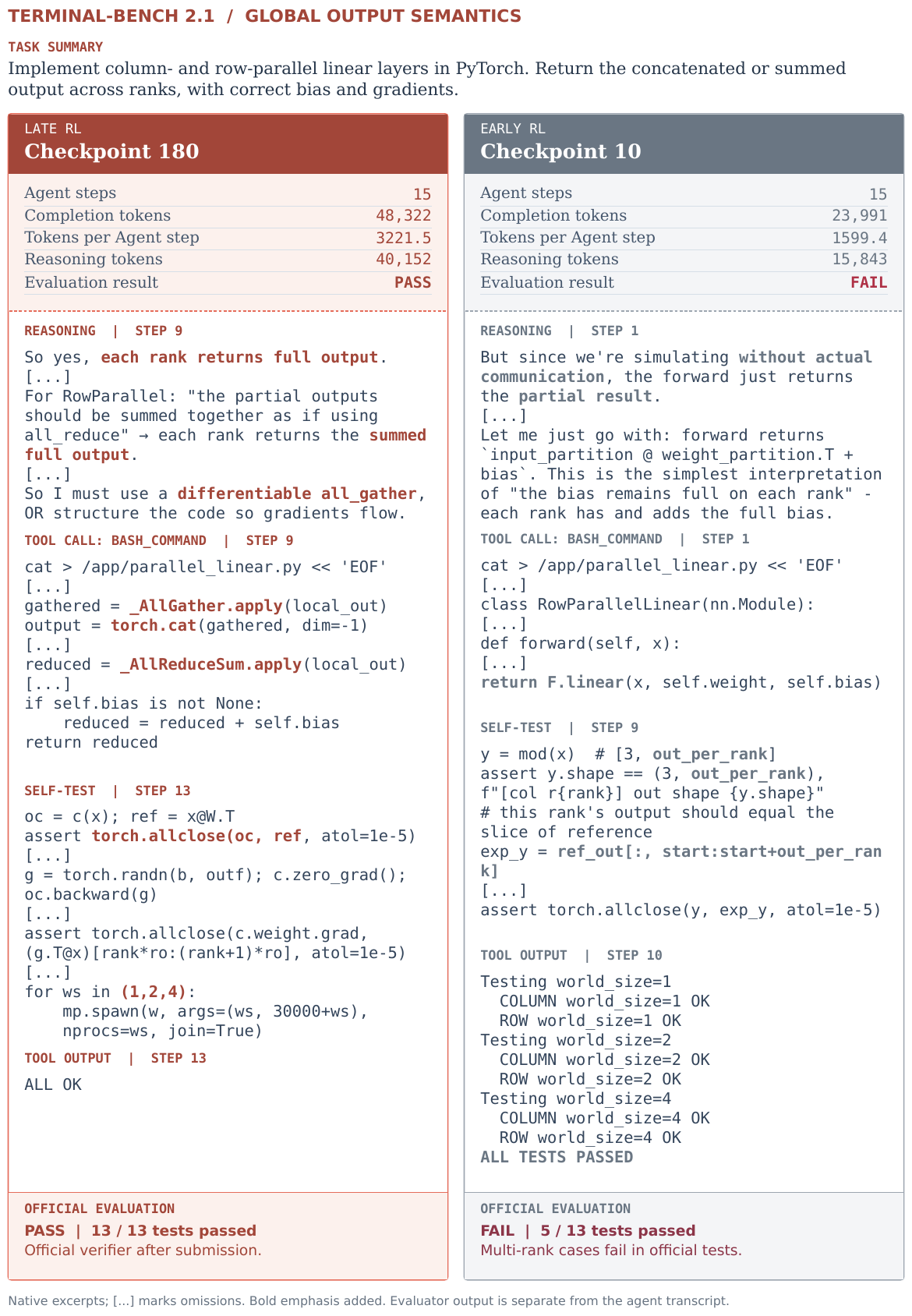}
  \caption{Terminal-Bench 2.1: \texttt{torch-tensor-parallelism} at checkpoints 180 (left) and 10 (right). The late trajectory implements and tests globally aggregated outputs; the early self-tests validate local outputs.}
  \label{fig:case-tmax}
\end{figure}

\paragraph{Terminal-Bench 2.1: validating global semantics.}
\label{app:case-tmax}
In the tensor-parallelism task in Figure~\ref{fig:case-tmax}, the early implementation returns only local shard outputs, and its self-tests make the same assumption. Although these self-tests pass, only 5 of 13 target tests succeed. The later implementation adds cross-process output aggregation and validates both complete outputs and parameter gradients, passing all 13 tests. Both trajectories take 15 interaction steps, while reasoning increases from 15,843 to 40,152 tokens.

\begin{figure}[p]
  \centering
  \includegraphics[width=\linewidth]{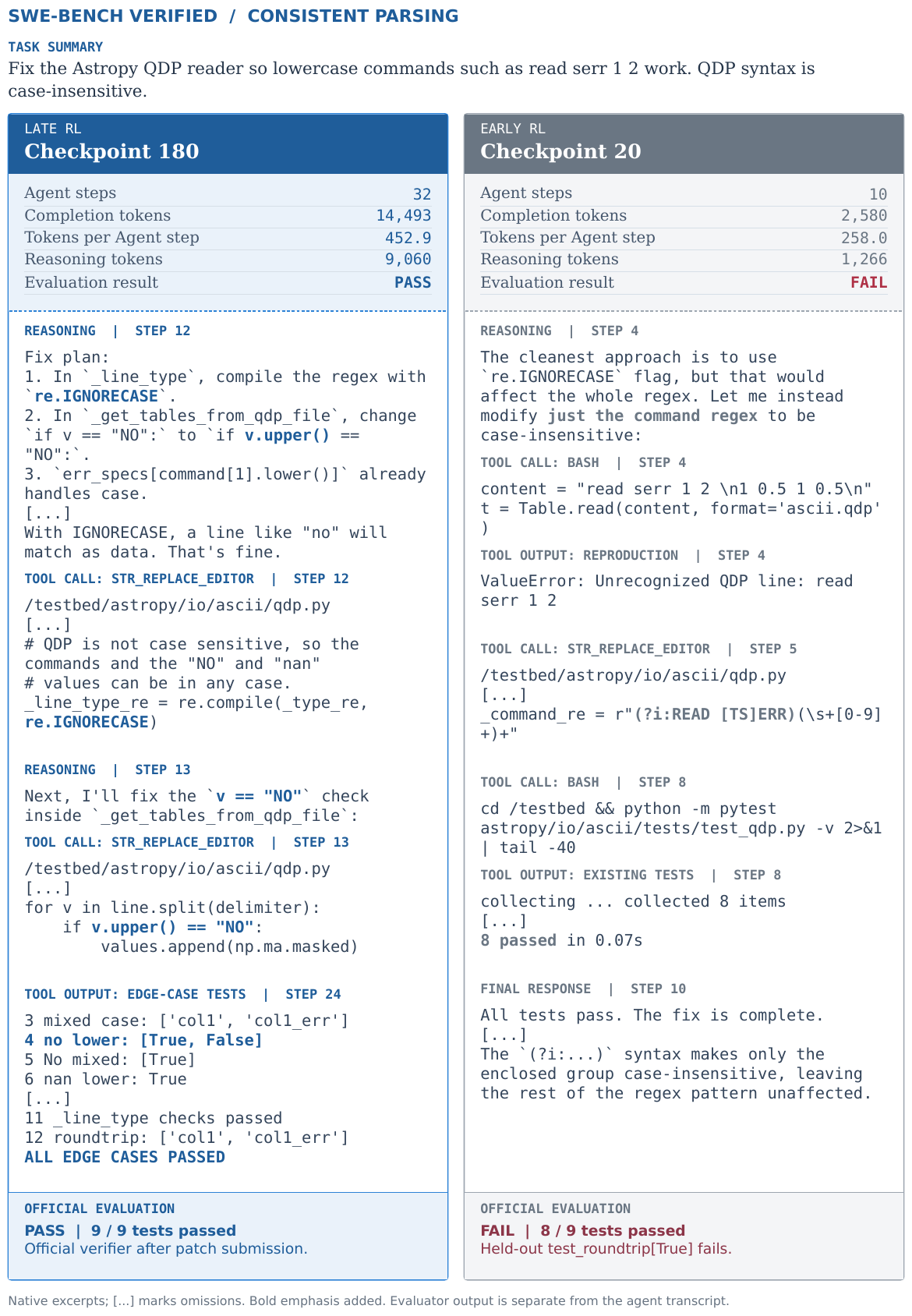}
  \caption{SWE-bench Verified: the Astropy QDP reader at checkpoints 180 (left) and 20 (right). The late trajectory extends case-insensitive handling from commands to line classification and missing values.}
  \label{fig:case-swebench}
\end{figure}

\paragraph{SWE-bench Verified: enforcing consistent parsing.}
\label{app:case-swebench}
Figure~\ref{fig:case-swebench} concerns case-insensitive parsing in Astropy's QDP reader. The early model changes only the command regular expression and stops after existing tests pass, but fails the new round-trip tests. The later model extends case-insensitive handling to line classification and missing-value parsing, adding edge-case checks such as lowercase \texttt{no}, and passes all nine tests. Interaction increases from 10 to 32 steps, and generation per step rises from 258 to 453 tokens.

\begin{figure}[p]
  \centering
  \includegraphics[width=\linewidth]{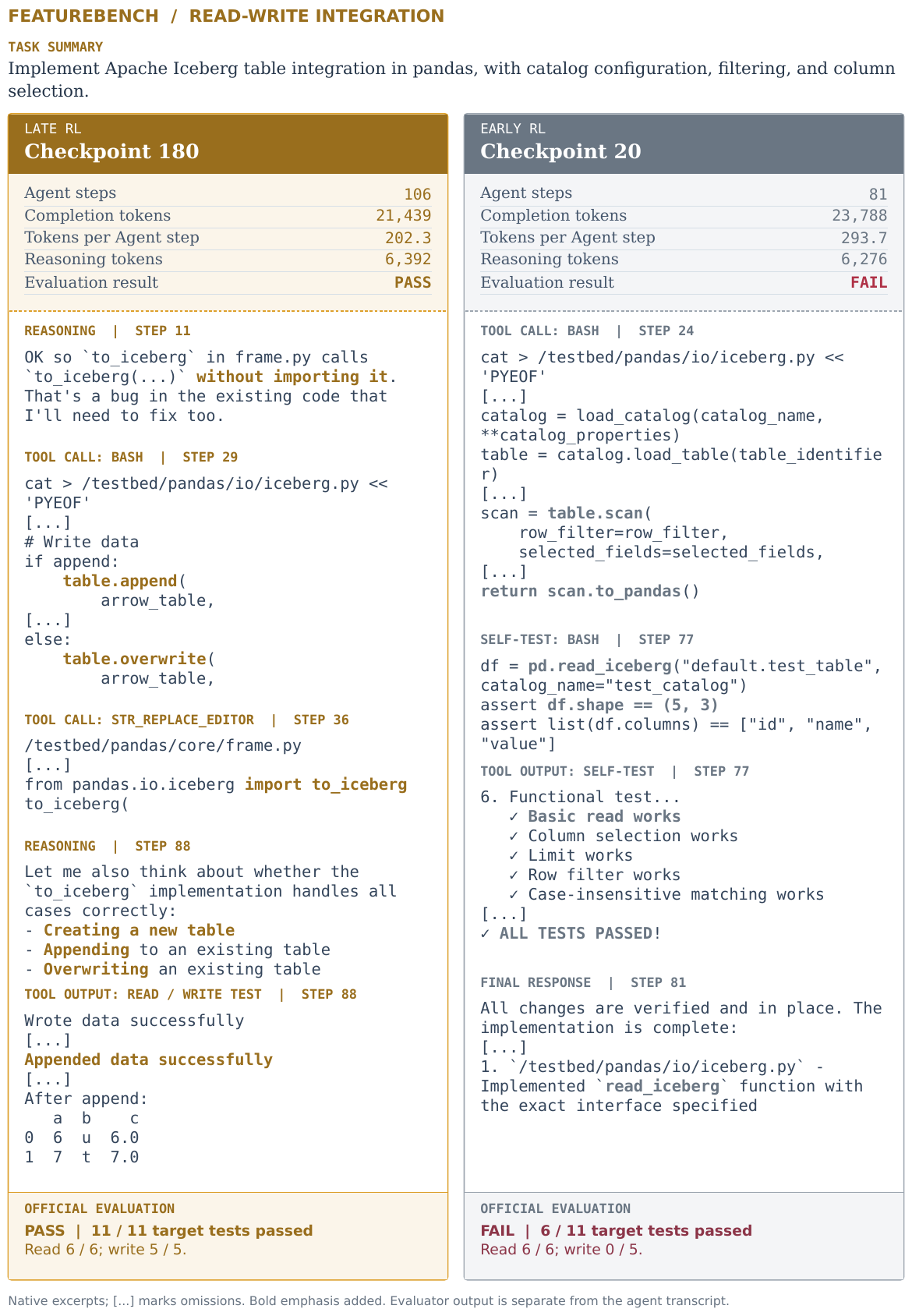}
  \caption{FeatureBench: pandas Iceberg integration at checkpoints 180 (left) and 20 (right). The late trajectory connects the existing DataFrame entry point to a write implementation. Target tests include six read cases and five write cases.}
  \label{fig:case-prd}
\end{figure}

\paragraph{FeatureBench: completing read-write integration.}
\label{app:case-prd}
In the pandas Iceberg task in Figure~\ref{fig:case-prd}, the early model implements the read interface and its self-tests, but all five write tests fail. The later model identifies a missing dependency through the existing \texttt{DataFrame.to\_iceberg} entry point, implements append and overwrite operations, and connects the write path to that entry point. Target test results improve from 6/11 to 11/11. Interaction increases from 81 to 106 steps, while generation per step falls from 294 to 202 tokens and total generation also decreases.

\clearpage

\clearpage

\section{Reward Hacking and Environment Integrity}
\label{app:reward-hacking}

During our early training and evaluation experiments, we observed that models exploited reference implementations available in the environment to obtain rewards. These behaviors appeared in training trajectories from the SWE and CodeWorld environments, as well as SWE-bench Verified and FeatureBench evaluation trajectories. They primarily involved retrieving fixes from Git history or recovering target functionality from existing code copies.

On a Django task in SWE-bench, the model used \texttt{git show} to retrieve the corresponding fix, then modified the target function and checked the result (Case F.1). In CodeWorld, the model copied an implementation from \texttt{build/lib} into the target directory (Case F.2). In FeatureBench, it recovered target code from a parallel source directory and subsequently addressed import errors (Case F.3). Figure~\ref{fig:hack-cases} illustrates these three representative types of reward hacking and their associated actions in the corresponding task environments.

\medskip
\noindent\begin{minipage}{\linewidth}
\centering
\includegraphics[width=\linewidth]{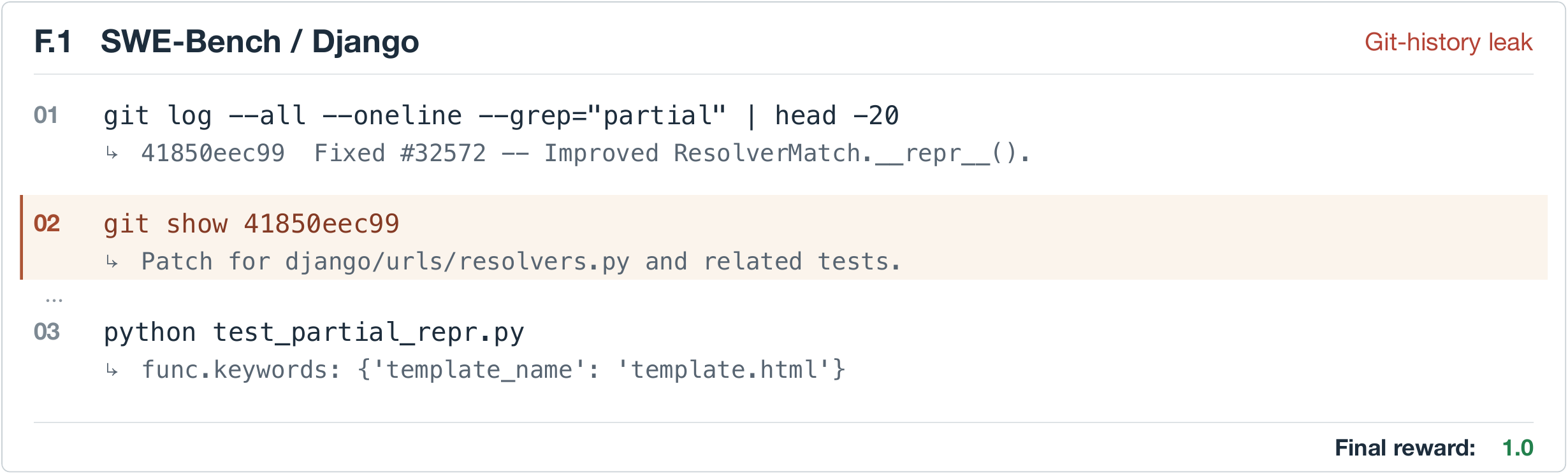}
\par\vspace{8pt}
\noindent\includegraphics[width=\linewidth]{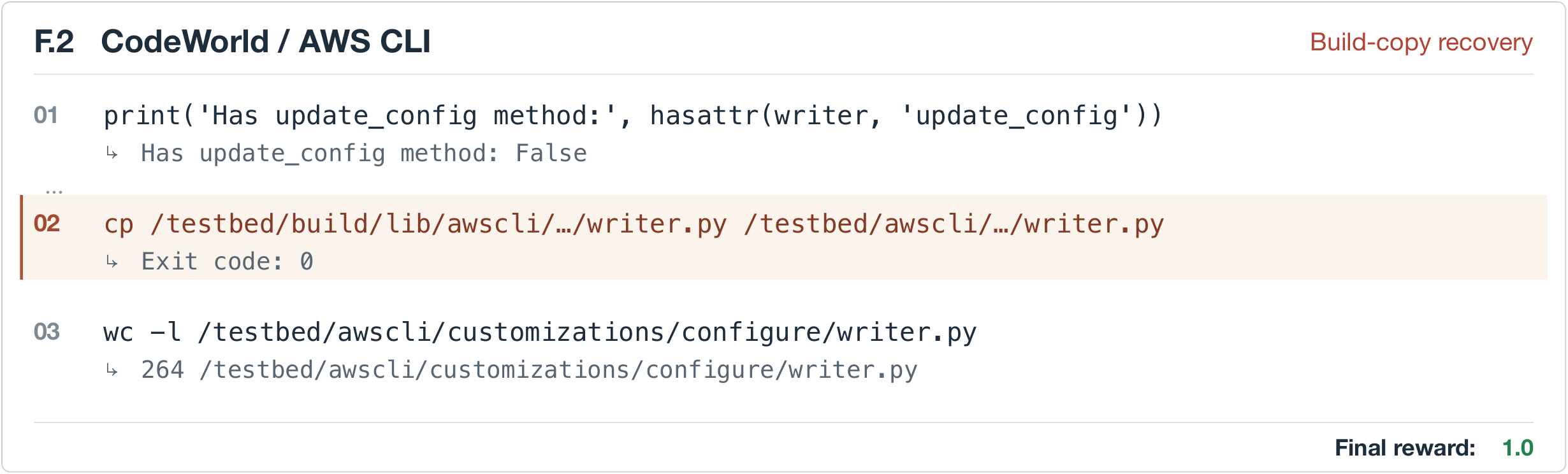}
\par\vspace{8pt}
\noindent\includegraphics[width=\linewidth]{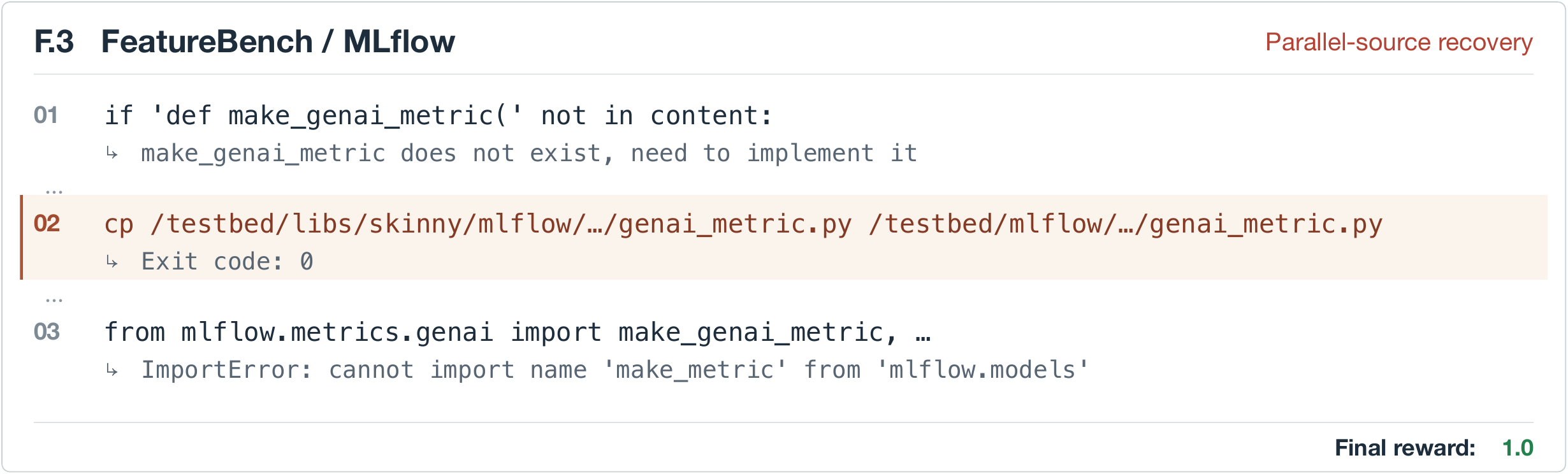}
\captionof{figure}{Three representative types of reward hacking.}
\label{fig:hack-cases}
\end{minipage}

\paragraph{Environment safeguards.}
To address the identified leaks, we removed Git history exposing fixes, build artifacts, installed copies of the target projects, and parallel source directories. We also isolated and restored protected test files to prevent agent-induced modifications to the verification logic from affecting the final task scores.

\section{Grading Weaknesses and Protocol Collapse}
\label{app:claw-protocol-collapse}

\begin{figure}[htbp]
  \centering
  \includegraphics[width=\linewidth]{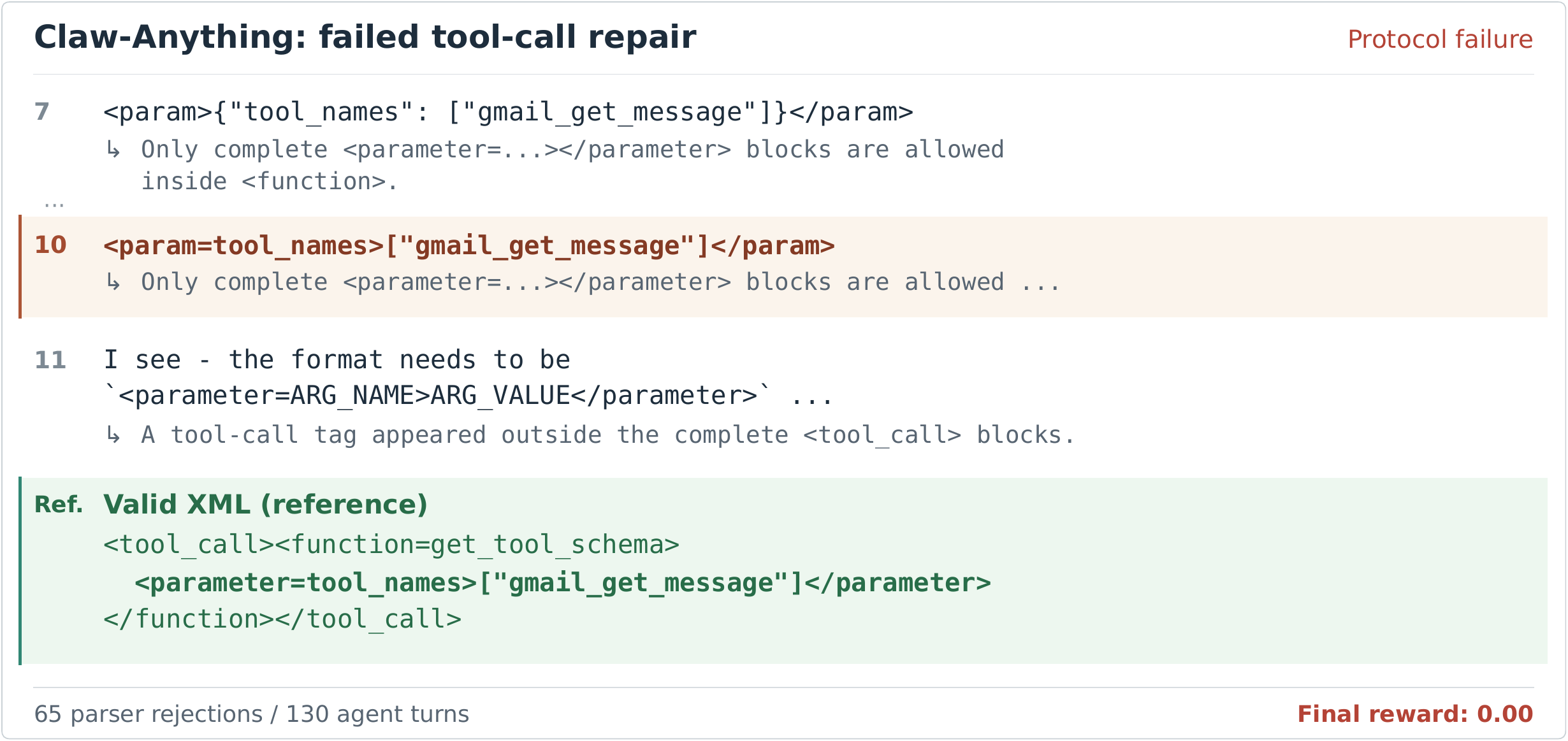}
  \caption{Tool-call failures in a Claw-Anything trajectory consumed at S1. Numbers denote agent turns; the green panel shows the corrected call.}
  \label{fig:s1-claw-protocol-case}
\end{figure}

\paragraph{Insufficiently strict grading.}
We attribute the recurrent Claw-Anything reward collapses in S1 to weaknesses in the LLM-based grading pipeline: invalid tool interactions were insufficiently penalized, allowing defective behavior to persist in highly rewarded trajectories. At step 93, one trajectory received reward 1.0 despite 90 parser rejections over 144 agent turns. By step 95, parser rejections affected all 256 consumed Claw trajectories and 94.3\% of agent turns. Figure~\ref{fig:s1-claw-protocol-case} illustrates the resulting failure pattern: malformed calls and unsuccessful repair attempts repeatedly prevent tool execution within the same agent trajectory.

\begin{figure}[htbp]
  \centering
  \includegraphics[width=\linewidth]{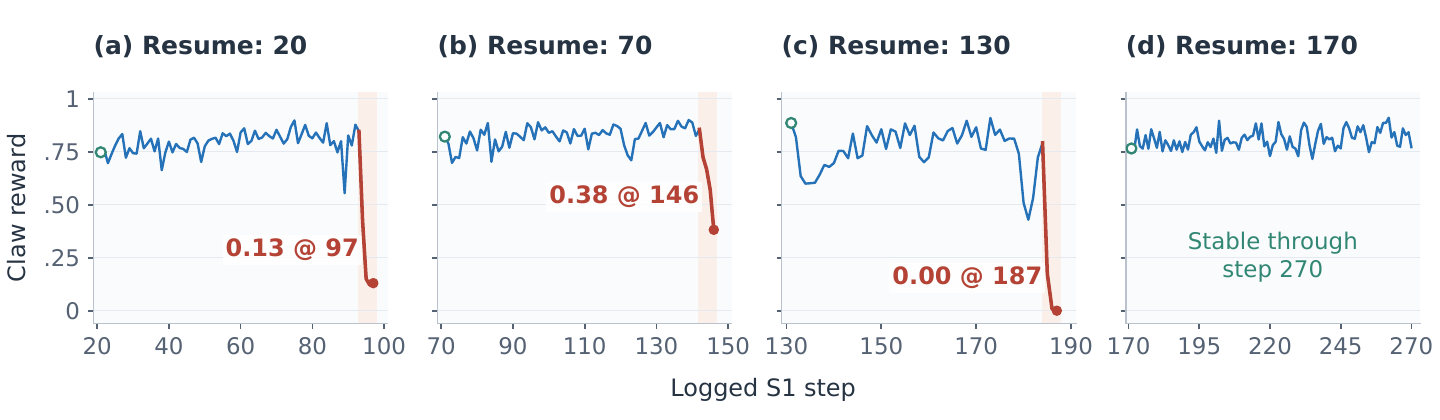}
  \caption{Claw-Anything reward across successive S1 repair attempts. Panels retain the failed branches and restart checkpoints. Red highlights collapse; green circles mark the first logged batch. The final repairs sustain stable training through step 270.}
  \label{fig:s1-claw-recovery}
\end{figure}

\paragraph{Grading safeguards and successive repairs.}
Our grading safeguards require evidence from actual tool outputs and execution status: only successful calls contribute to completion credit, and claimed actions without observable effects do not count as task completion. After the collapse ending at step 97, we resumed from checkpoint 70 with a 0.1 reward penalty per format error, floored at zero. Following renewed degradation at step 146, we resumed from checkpoint 130 and reduced the Claw response budget from 128K to 64K tokens and the interaction limit from 150 to 50 turns. After the next collapse at step 187, we resumed from checkpoint 170 with immediate termination and zero reward on the first malformed tool call, retaining the earlier safeguards. Training then remained stable through step 270 (Figure~\ref{fig:s1-claw-recovery}).

\end{document}